\documentclass[10pt,twocolumn,letterpaper]{article}

\usepackage{cvpr}
\usepackage{times}
\usepackage{epsfig}
\usepackage{graphicx}
\usepackage{amsmath}
\usepackage{amssymb}
\usepackage{subcaption}
\DeclareMathOperator*{\argmax}{arg\,max}

\usepackage[pagebackref=true,breaklinks=true,letterpaper=true,colorlinks,bookmarks=false]{hyperref}

\cvprfinalcopy 

\def\cvprPaperID{2709} 

\ifcvprfinal\fi
\begin{document}

\title{Learning Pixel Representations for Generic Segmentation}

\author{Oran Shayer\\
Technion - Israel Institute of Technology\\
Haifa, Israel\\
{\tt\small oran.sh@gmail.com}
\and
Michael Lindenbaum\\
Technion - Israel Institute of Technology\\
Haifa, Israel\\
{\tt\small mic@cs.technion.ac.il}
}

\maketitle

\begin{abstract}
Deep learning approaches to generic (non-semantic) segmentation have so far been indirect and relied on edge detection. This is in contrast to semantic segmentation, where DNNs are applied directly. We propose an alternative approach called Deep Generic Segmentation (DGS) and try to follow the path used for semantic segmentation.
Our main contribution is a new method for learning a pixel-wise representation that reflects segment relatedness. This representation is combined with a CRF to yield the segmentation algorithm.
We show that we are able to learn meaningful representations that improve segmentation quality and that the representations themselves achieve state-of-the-art segment similarity scores. The segmentation results are competitive and promising. 

\end{abstract}

\section{Introduction}

Generic segmentation is the well-studied task of partitioning an image into parts that correspond to objects for which no prior information is available. Deep learning approaches to this task thus far have been indirect and relied on a high quality edge detector. The COB algorithm \cite{cob} for example, produces high quality segmentations by learning an oriented contour map and creating segmentation hierarchies using the oriented watershed transform.

In this work we consider an alternative approach that does not rely on edge detectors, but rather follows the approach used for semantic segmentation: learning pixel-wise representations that capture segment characteristics and help generate meaningful segmentations. This paper focuses on creating such representations, along with an initial attempt to apply them.

Deep learning has been successfully used in a supervised regime, where the network is learned end-to-end on a supervised task (classification \cite{resnet}, object detection \cite{yolo}, semantic segmentation \cite{deeplab} or edge detection \cite{hed}). Generic segmentation, however, cannot be formulated as such. Segments in new (test) images are not well specified with respect to the segments or the objects in a training set, and therefore the direct classification approach does not apply.

The common task of face verification (\cite{deepface}, \cite{facenet}, \cite{face_veldadi}) shares this difficulty. Even if we learn on thousands of labeled faces, there may be  millions of unseen faces that must be handled. To succeed in this task, we must be able to learn a model, or representation, that can capture properties and characteristics capable of distinguishing between different classes, even those not encountered in training.
Similarly, in generic segmentation, the examples to be partitioned might contain objects not seen in the training set. The problem is further complicated by two factors: the annotated segments have unknown semantic meaning (i.e., we do not know what objects or parts are marked), and the number of segments in each image is also unknown.

Face verification is made using the relations between the face representations in feature space and we follow this approach. We therefore aim to learn pixel-wise representations that express segment relatedness. We learn representations that are grouped together in representation space for pixels of the same segment, and kept further apart for pixels from different segments.
In this paper, we propose a novel approach to learn such pixel-wise representations. We compare it to the more common approach for deep representation learning (triplet loss \cite{hoffer2015triplet}), and test it quantitatively and visually. We show that the best way to learn such representations is by a new supervised algorithm which follows the principles of the DeepFace algorithm \cite{deepface} but addresses the differences between segmentation and face classification.

\begin{figure}[t]
\centering
\includegraphics[width=0.48\textwidth]{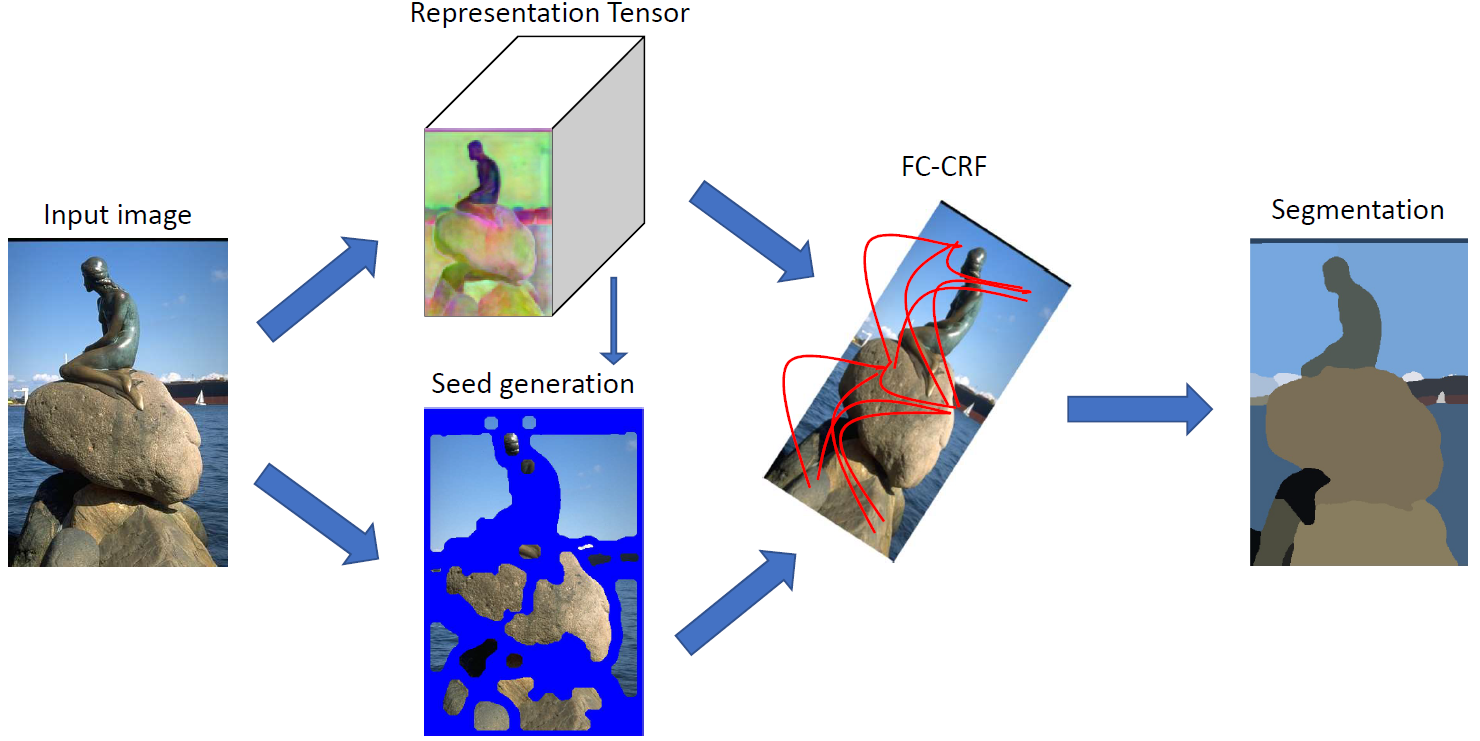}
\caption{Our overall pipeline. We extract the representations and generate seed regions from which we set the unary values of a FC-CRF. Performing the estimated inference over the CRF will output the final segmentation.}\label{fig:pipeline}
\end{figure}

Our deep generic segmentation (DGS) algorithm uses the learned representation for seed generation and for CRF inference, as illustrated in Fig. \ref{fig:pipeline}.
%
%
%
Our contributions in this paper are as follows:
\begin{enumerate}
\item We present (the first) pixel-wise representations tailored specifically for generic segmentation. These representations capture segmentation properties and perform better than previous methods on a pixel pair classification task.
\item We implement a new deep learning approach for learning such pixel-wise segment representations.
\item We present a new segmentation algorithm that makes use of these representations and further demonstrate their effectiveness. 
\end{enumerate}

\section{Related Work}

\subsection{Generic segmentation}
Generic segmentation has seen a broad range of approaches and methods. Here we mention only a few examples. Earlier methods such as the mean shift algorithm \cite{meanshift} rely on clustering of local features.
%
%

Graph representations are commonly used for segmentation algorithms. There, pixels or other image elements are represented by graph nodes, and weights on the edges represent the similarity between them. The intuitive idea of dividing the image into two parts that are most dissimilar translates into finding the minimal cut in this similarity graph. 
The Normalized Cuts \cite{ncuts} criterion  guarantees that no group is too small, and the problem of finding the minimal normalized cut is elegantly solved using generalized eigenvectors. 
The weights on the edges can also represent the dissimilarity between the nodes. The bottom-up watershed algorithm \cite{watershed} merges at each iteration the two elements with the smallest dissimilarity values between them. A merging criterion that is sensitive to the typical dissimilarity between the merged segments and to their sizes significantly improves the results  \cite{felzhutt}.

The OWT-UCM algorithm \cite{ucm} uses an (oriented) edge detector to get reliable dissimilarities, and transforms the graph into a hierarchical region tree. It then modifies the weights so that thresholding them yields a set of closed curves and well-specified segmentation. Using different thresholds gives the hierarchy. Combining multiple scales further improves performance \cite{mcg}. This approach, coupled with a CNN edge detector, achieves the current state-of-the-art \cite{cob}.
%
%


\subsection{Semantic segmentation}
Deep semantic segmentation builds on the ability of fully convolutional NNs to identify the pixels associated with particular categories \cite{fcn}. Skip connections, deconvolution, and dilated (atrous) convolutions were used to maintain and improve output resolution \cite{fcn,deconvnet,deeplab}. Pyramid pooling \cite{psp} and encoder-decoder architecture \cite{large_kernel_matters} were used to capture larger context. A combination with CRF was used to get better localized boundaries\cite{fc-crf}. 

\subsection{Representation learning}

Representations can be learned explicitly using a \textit{Siamese network} (\cite{siamese2005, koch15, siamese2006}). An example is a pair of inputs, either tagged as same (positive example) or not same (negative example). Both inputs are mapped to a representation through neural networks whose weights are tied. The networks are then trained to minimize the distance in representation space between positive examples and increase the distance between negative examples.
An example may also be a triplet of inputs (\cite{hoffer2015triplet,facenet}), where the first two inputs are positive and negative, respectively, relative to the third one.

The representation can also be learned implicitly by learning a supervised task. In a straightforward supervised setting, the last layer of the network can be regarded as a classifier, and the rest of the network can be regarded as generating a representation that will be fed to this classifier (\cite{decaf}). These representations can be used later on to distinguish between unseen classes \cite{deepface}, or for transfer learning \cite{decaf}.

The closest work to ours is Patch2Vec \cite{p2v}, which (explicitly) learns an embedding for image patches by training on triplets tagged according to the segmentation. Our approach differs in that it uses implicit learning and allows a larger context available from the full image.

\section{Representations for Non-semantic Segmentation}
\label{representation}

One well-known strength of neural networks is their ability to capture both low level and high level features of images, creating powerful and useful representations (\cite{zeiler2014visualizing, bengio2013representation}). 
In this work, we focus on segmentation-related representations and harness this strength to provide a new pixel-wise representation. This learned representation should capture the segment properties of each pixel, so that representations associated with pixels that belong to the same segment are close in representation space, and their cluster is farther away from clusters representing different segments.
This representation is a pixel-wise $N$-dimensional vector (thus, the full image representation, denoted $R$, is an $H\times W\times N$ tensor). 

Learned classifiers have been used in the context of semantic (model based) segmentation. Learning a classifier directly is possible for the semantic segmentation task because every pixel is associated with a clear label: either a specific category or background.  This is not the case with generic segmentation, where object category labels are not available during learning and are not important at inference. Moreover, at inference the categories associated with the segments are not necessarily those used in training.

\subsection{Learning the Representation}\label{Learning-rep}

\subsubsection{Explicit learning -- Siamese or triplet loss}\label{explicit}
A pixel-wise representation for generic segmentation can be learned directly by minimizing a Siamese loss function over same-not same pixel pairs (\cite{siamese2006}), or a triplet loss over triplets \cite{hoffer2015triplet,facenet}. For example, if we denote the same label of pixels $i$ and $j$ as $Y_{ij}=1$  and not same as $Y_{ij}=0$, then we can train in a Siamese setting
\begin{align}
  \mathcal{L}_{siamese} = \sum_{i,j} L(R_i, R_j, Y_{ij}),
\end{align}
where
\begin{align}
\begin{split}
L(R_i, R_j, Y_{ij}) & = Y_{ij}\left[d(R_i, R_j)^2\right] \\
& + (1-Y_{ij})\left[max(0, m-d(R_i, R_j)\right],
\end{split}
\end{align}
where $d(\cdot,\cdot)$ is some measure of difference and $m$ is some margin value . A similar approach can be taken with a triplet setting. A triplet input consists of an anchor, a positive example (same segment) with respect to the anchor, and a negative example (not same) relative to the anchor. We then minimize the following triplet loss:
\begin{align}
  \mathcal{L}_{triplet} = \sum_{i} L(R_i, R_j^+, R_k^-),
\end{align}
where
\begin{align}
L(R_i, R_j^+, R_k^-) = \left[d(R_i, R_j^+)\right]^2 + max\left[0, m-d(R_i, R_k^-)\right]
\end{align}
While this approach has been proved beneficial for high-level image representations or patches, it has not been explored on tasks which provide structured outputs such as segmentation.

\subsubsection{Implicit learning}\label{implicit}
\begin{figure}[!h]
\centering
\includegraphics[width=0.48\textwidth]{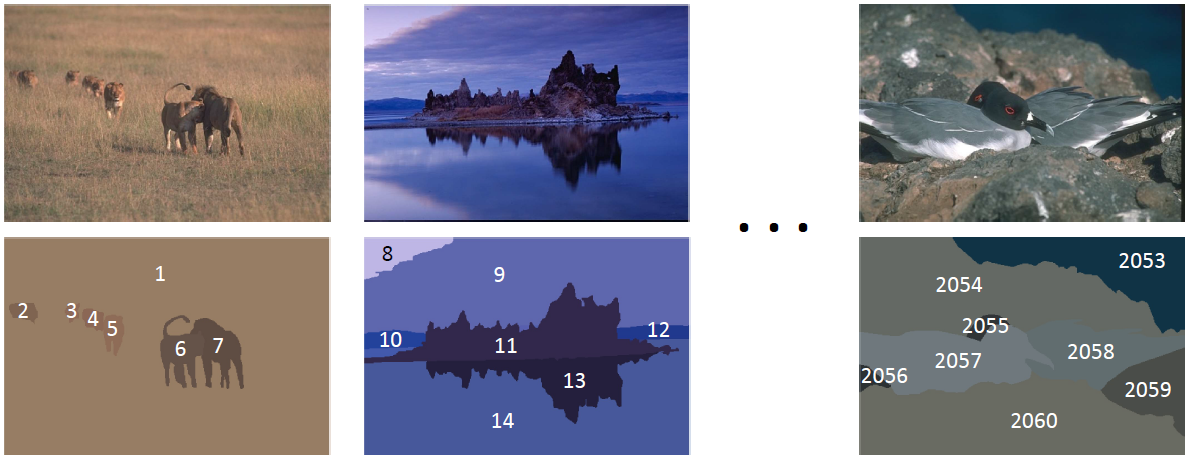}
\caption{Our labeling process. We assign a unique label $l_k$ to each segment in every training image.}\label{fig:labeling}
\end{figure}

\begin{figure*}[h]
\centering

\begin{subfigure}[b]{0.22\textwidth}
  \includegraphics[width=\textwidth]{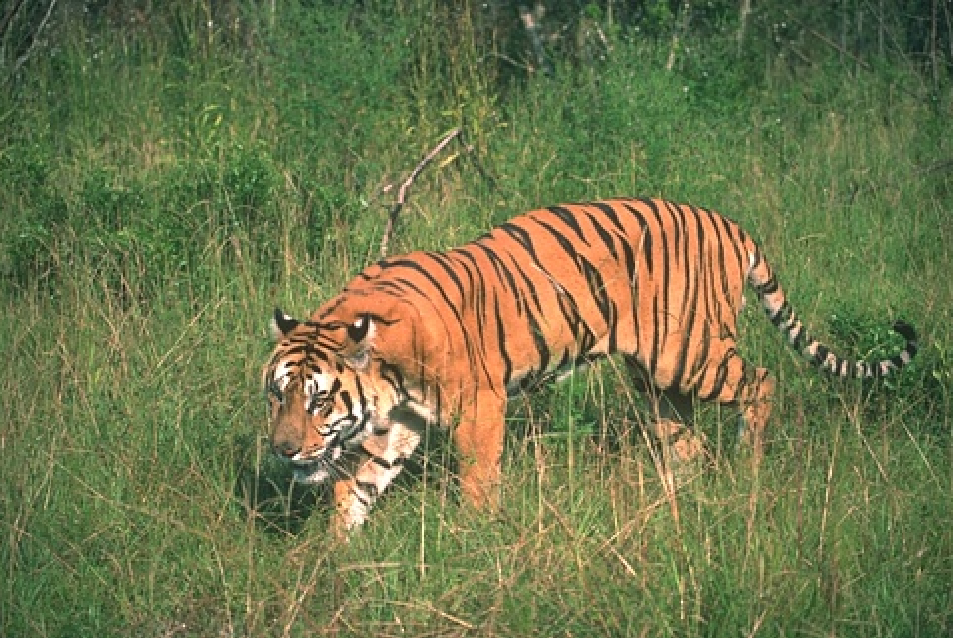}
\end{subfigure}
\begin{subfigure}[b]{0.22\textwidth}
  \includegraphics[width=\textwidth]{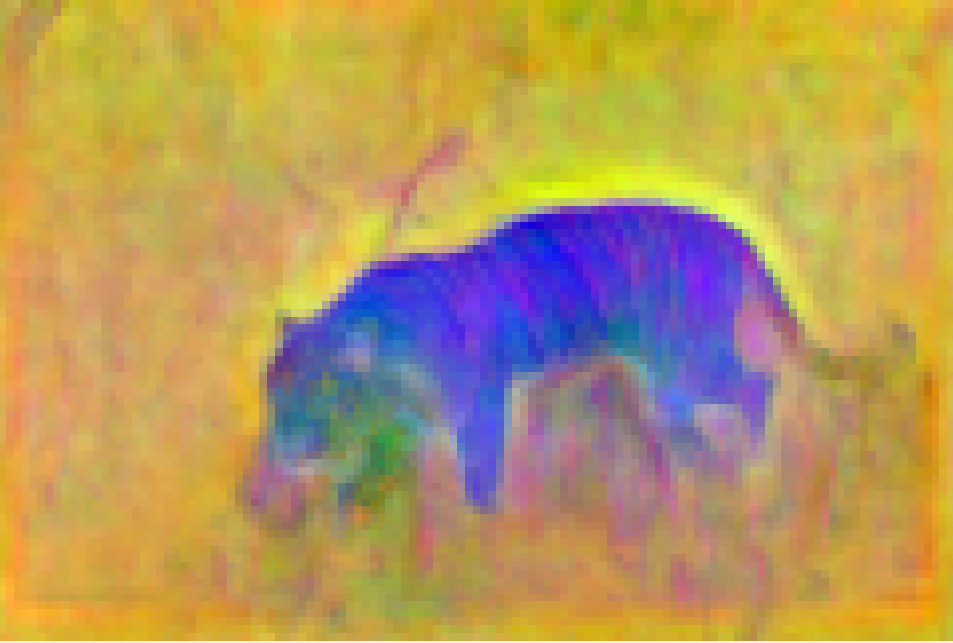}
\end{subfigure}
\begin{subfigure}[b]{0.19\textwidth}
  \includegraphics[width=\textwidth]{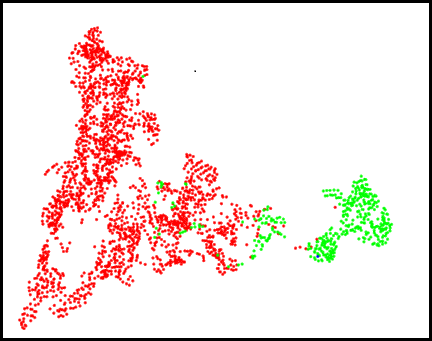}
\end{subfigure}
\begin{subfigure}[b]{0.19\textwidth}
  \includegraphics[width=\textwidth]{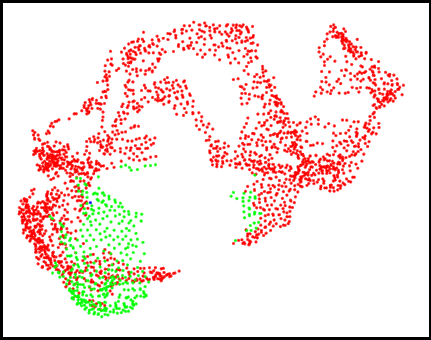}
\end{subfigure}

\begin{subfigure}[b]{0.22\textwidth}
  \includegraphics[width=\textwidth]{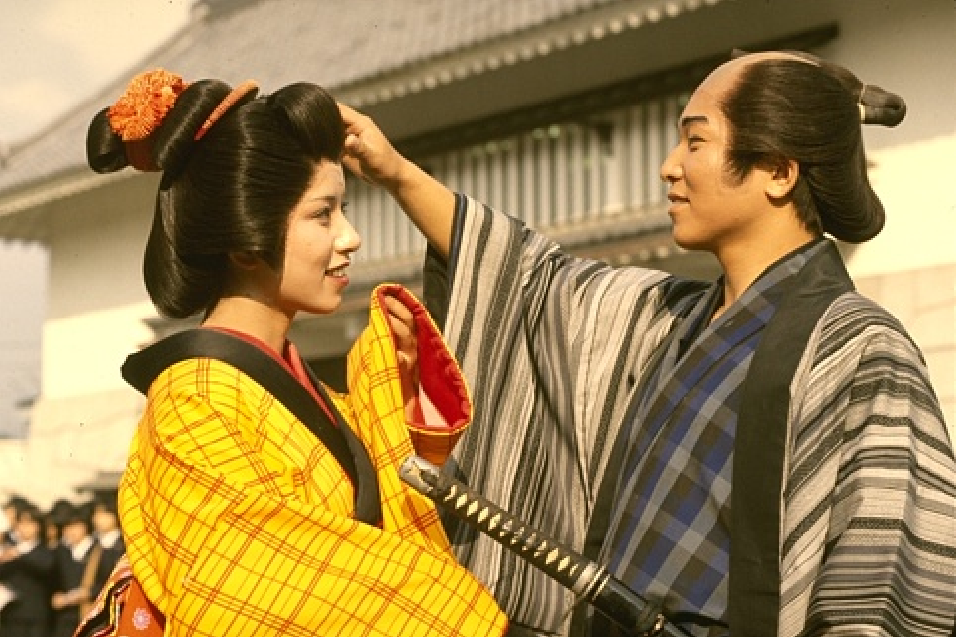}
\end{subfigure}
\begin{subfigure}[b]{0.22\textwidth}
  \includegraphics[width=\textwidth]{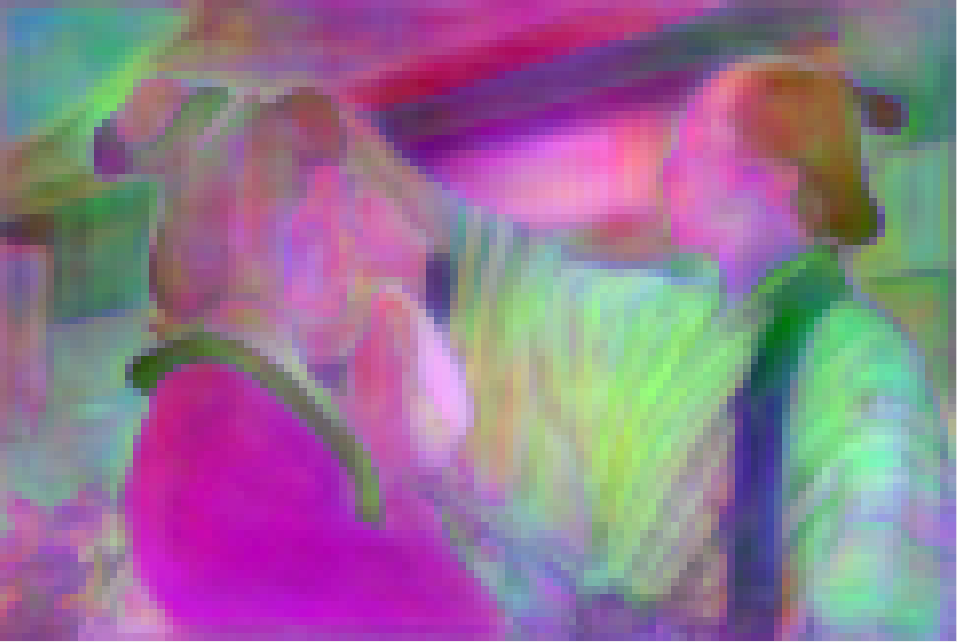}
\end{subfigure}
\begin{subfigure}[b]{0.19\textwidth}
  \includegraphics[width=\textwidth]{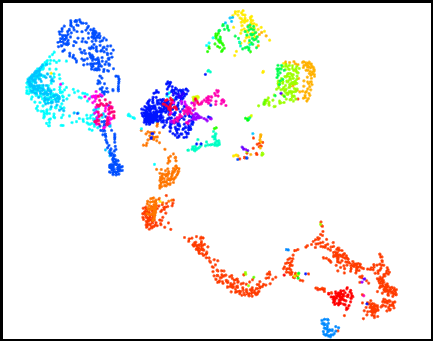}
\end{subfigure}
\begin{subfigure}[b]{0.19\textwidth}
  \includegraphics[width=\textwidth]{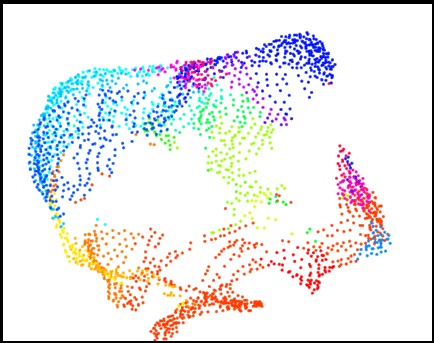}
\end{subfigure}

\begin{subfigure}[b]{0.22\textwidth}
  \includegraphics[width=\textwidth]{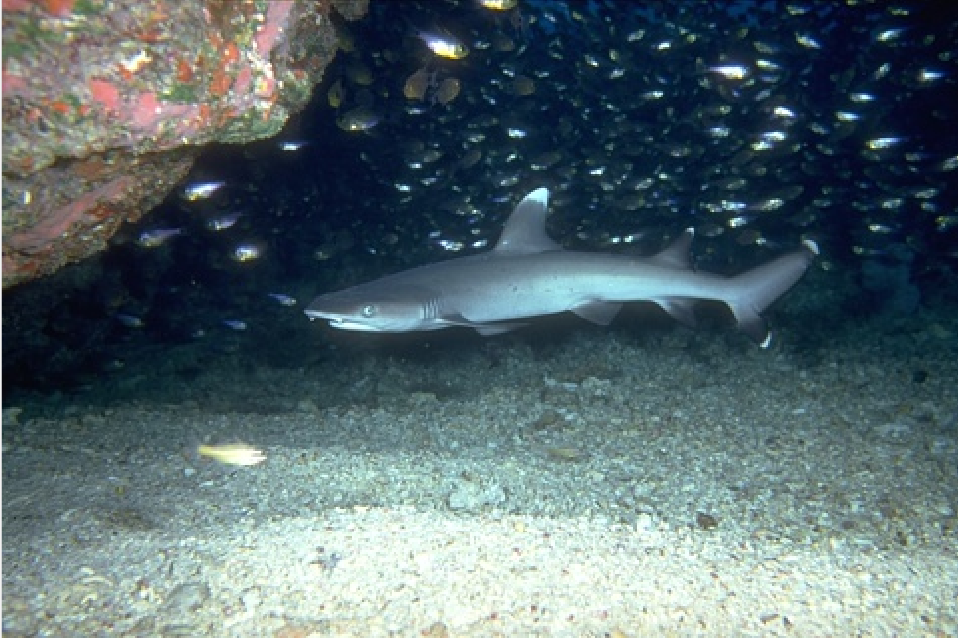}
  \caption{}
\end{subfigure}
\begin{subfigure}[b]{0.22\textwidth}
  \includegraphics[width=\textwidth]{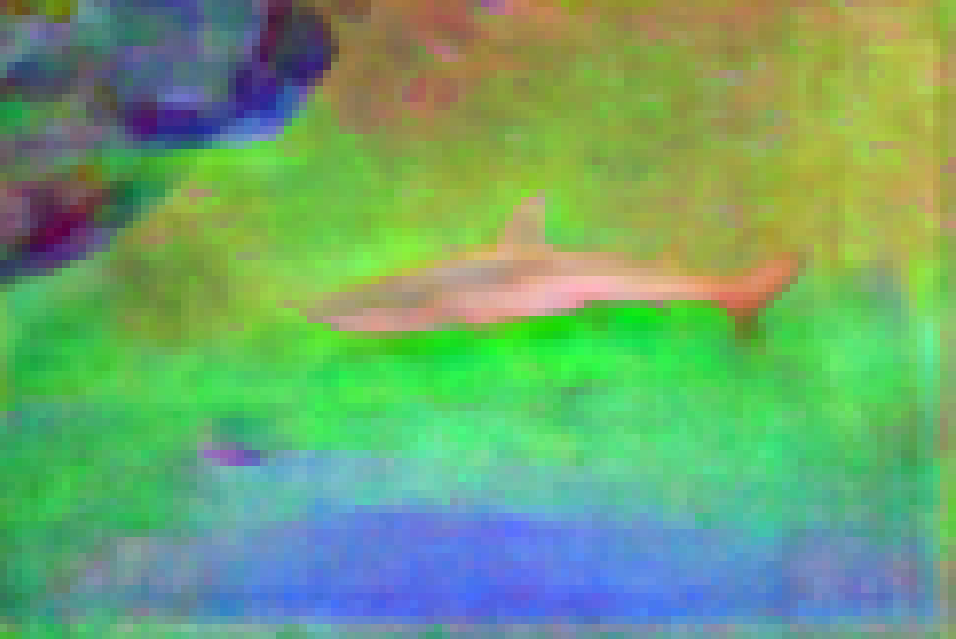}
  \caption{}\label{fig:reps-pca}
\end{subfigure}
\begin{subfigure}[b]{0.19\textwidth}
  \includegraphics[width=\textwidth]{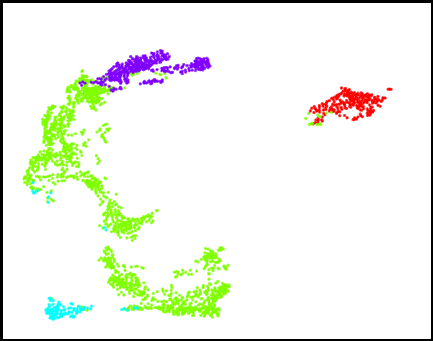}
  \caption{}
\end{subfigure}
\begin{subfigure}[b]{0.19\textwidth}
  \includegraphics[width=\textwidth]{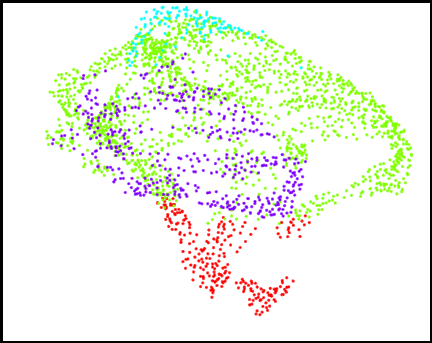}
  \caption{}
\end{subfigure}

\caption{visualizations of our proposed representation (sec. \ref{implicit}). In each row, from left to right: (a) the original image, (b) the representation space virtual colors, (c) t-SNE of our representation and (d) t-SNE of the representation from a network trained for semantic segmentation. In the t-SNE plots points of the same color belong to the same segment. Notice how areas such as the tiger in the first image or the woman's kimono in the second image are nearly uniform in color, indicating that those pixels are close in representation space. In addition, notice the sharp boundaries in the virtual colors, evidence that the representation is boundary aware and represents the pixel accordingly. The t-SNE plots also show that clusters formed from our representation are much more promising.}\label{pca_tsne}
\end{figure*}

Representations can also be learned implicitly, by training a network on a related high-level task, and afterwards using the representation from the last hidden layer for the original task. A related task is face recognition in the wild, or face verification, where the categories (face identities) to be classified at test time are usually different and more abundant than those available for training. In the DeepFace approach \cite{deepface}, the training algorithm learns a face classification task using examples of $K$ (over 4000)  face identity classes.
An $L$-layer classification network is trained, where the $L$th layer is the final linear classification layer.
The $N$-dimensional response  from layer $L-1$, denoted as $f$ in the original work, is used as the representation of the input image. The training criterion (cross-entropy) inherently forms clusters of face images belonging to the same identity \cite{decaf}.

By training over a sufficiently large number of face images, the network generalizes well and generates well-clustered representations even for new images of unseen categories. These representations were used successfully to separate and classify new unseen faces. This process is reminiscent of our generic segmentation task where the aim is to separate pixels that belong to different segments not seen in training. We propose to adopt this approach for learning a representation for generic segmentation. However, some differences need to be addressed. First, here we are interested in a representation for every pixel and not for the full image. A fully convolutional network would output a tensor where every output pixel is represented by some $N$-dimensional vector.

A more fundamental difference is that choosing the training labels  is not straightforward.  Pixels in segmentation examples are assigned labels depending on the segment they belong to, but unlike face identities, the labels associated with different segments are not meaningful in the sense that they are not associated with object categories or even with appearance types. Segments in different images, for example, may correspond to the same object category (e.g. a horse), but this information is not available for training. To address this problem, we consider the set of segments from all images in the training set as different categories. That is, we assign a unique label $l_{k}$ to all pixels in the  i-th segment of the $j$th image ($s_{ij}$), a label that no other pixel in another segment or image is assigned. A visualization of the labeling process can be seen in Fig. \ref{fig:labeling}.

The use of arbitrary categories leads, however, to several difficulties. Two segments of different images may correspond to the same object category and may be very similar (e.g. two segments containing blue sky) but they are considered to be of different classes. Because the two segments have essentially the same characteristics, training a network to discriminate between them would lead to representations that rely on small differences in their properties or on arbitrary properties (e.g. location in the image), both leading to poor generalization. To overcome this difficulty, we modify the training process: when training on a particular image, we  limit the possible predicted classes only to those that  correspond to segments in this image, and not to the segments in the entire training set. 

The fully convolutional NN is trained as a standard pixel-wise classification task which, when successful, will classify each pixel to the label of its segment. In that case, the network will have learned representations which are well-clustered for pixels in the same segment, and different for pixels in different segments. We denote the representation of the $i$th pixel by  $R_i$. The distance $||R_i - R_j||$ between two pixels should reflect the segment relatedness.

\subsection{Evaluating the representations}\label{validation-reps}

We compared the representation obtained from our proposed implicit learning method (sec. \ref{implicit}) with the triplet loss approach as well as with several other possible representations on a pixel classification task.
The task is to determine whether two pixels belong to the same segment using the representations. To this end we use a simple classifier which decides that the pixels belong to the same segment if the Euclidean distance between the representations is smaller than a threshold. The optimal threshold will be learned over a validation set. The classification results of the test set are presented in sec. \ref{rep-results}.

\subsection{Visualizing the representations}

We suggest two options for visualizing the representations.
\begin{enumerate}
\item \textbf{Representation space virtual colors --} We project the $N$-dimensional representations on their three principal components (calculated using the PCA of all representations). The three-dimensional vector of projections is visualized as an RGB image. We expect pixels in the same segment to have similar projections and similar color. Fig. \ref{pca_tsne} shows this is indeed the case.
\item 
\textbf{t-SNE scatter diagram --} Intuition about the representations can also be gained by using t-Distributed Stochastic Neighbor Embedding (t-SNE, \cite{t-SNE}) to embed them in a 2D space. We expect a separation between points belonging to different segments in the embedded space, and compare the separation to that obtained by a network with the same architecture but trained for semantic segmentation. See  Fig. \ref{pca_tsne}.
\end{enumerate}
Note that the representation changes sharply along boundaries (Fig. \ref{fig:reps-pca}). This behavior is typical to nonlinear filters (\eg bilateral filter) and therefore indicates that the representation at a pixel depends mostly on image values associated with the segment containing the pixel and not on image values at nearby pixels outside this segment. That is, the representation describes the segment and is not a simple texture description associated with a uniform neighborhood.

\section{A Segmentation Algorithm}
\label{algorithm}

The learned representation is used as a part of a segmentation algorithm we propose, called Deep Generic Segmentation (DGS). It takes a top-down approach, in contrast to the common  bottom-up approach (\cite{cob,mcg}) dominant in generic segmentation. That is, it starts by making hypotheses about the segments and continues by refining them. The algorithm consists of two main stages.

\begin{description}
\item[Seeds generation and merging - ] A set of seeds constitutes a hypothesis about the number of segments and their approximate locations. Unlike the common seed concept, here, every seed is not a point but is rather a set of pixels that belong to the corresponding segment. The seed generation process proposes a set of initial seeds and then merges some of them based on the representation similarities and geodesic distance.

\item[Using the representation and a CRF -]  A probability that the pixel belongs to the $i$th seed is constructed using the distance in the representation space and the geodesic distance from the $i$th seed to the pixel.
These constructed probabilities are then used as the unary term and smoothed using a CRF.
\end{description}

\subsection{Stage 1: Seed Generation - A Direct Approach}\label{alg1}

\subsubsection{Initial seed regions}\label{intial-seeds-direct}
To generate the seed regions, we propose to apply, directly, the distance transform (DT) on the boundaries. Ideally, all segment boundaries, and only those boundaries, are associated with a zero DT value. Therefore, the connected components (CCs) of $DT>\epsilon$ are the segments.

Instead of estimating the edges and then running a DT on the resulting edge image, we use a modified version of the {\em deep watershed transform}, introduced and used for instance segmentation in \cite{dwt}, where the authors estimate the DT directly using a DNN.
We apply the same technique here, but over the whole image. 
By estimating the distance to the nearest boundary, the trained network has the potential of being more sensitive to the context around the pixel (compared with edge detection); hence, directly predicting the DT should be more accurate.

\begin{figure}[!h]
\centering

\begin{subfigure}[b]{0.14\textwidth}
  \includegraphics[width=\textwidth]{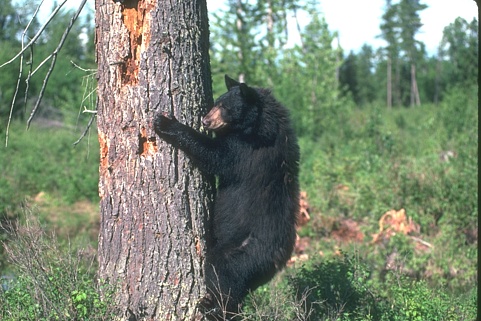}
  \caption{Image}
\end{subfigure}
\begin{subfigure}[b]{0.14\textwidth}
  \includegraphics[width=\textwidth]{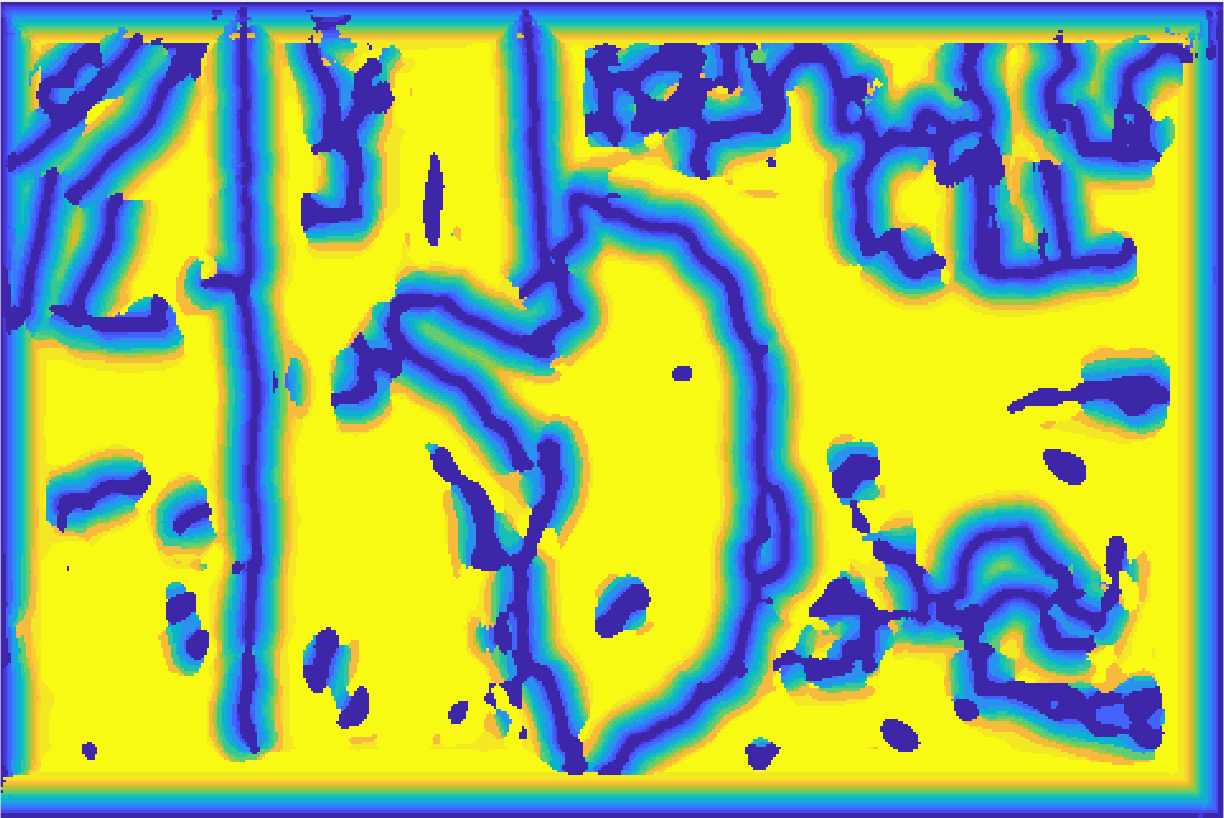}
  \caption{Estimated DT}
\end{subfigure}
\begin{subfigure}[b]{0.14\textwidth}
  \includegraphics[width=\textwidth]{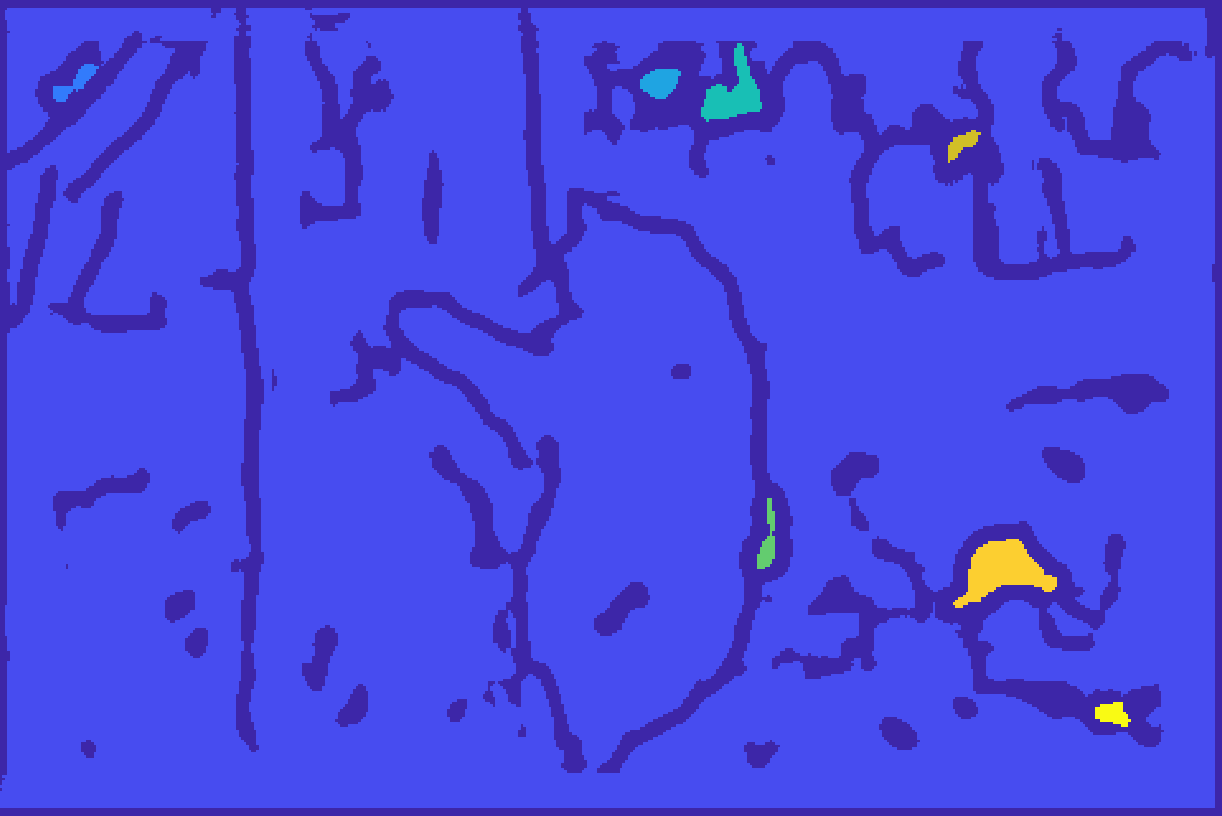}
  \caption{CCs, $DT>\epsilon$}
\end{subfigure}

\begin{subfigure}[b]{0.14\textwidth}
  \includegraphics[width=\textwidth]{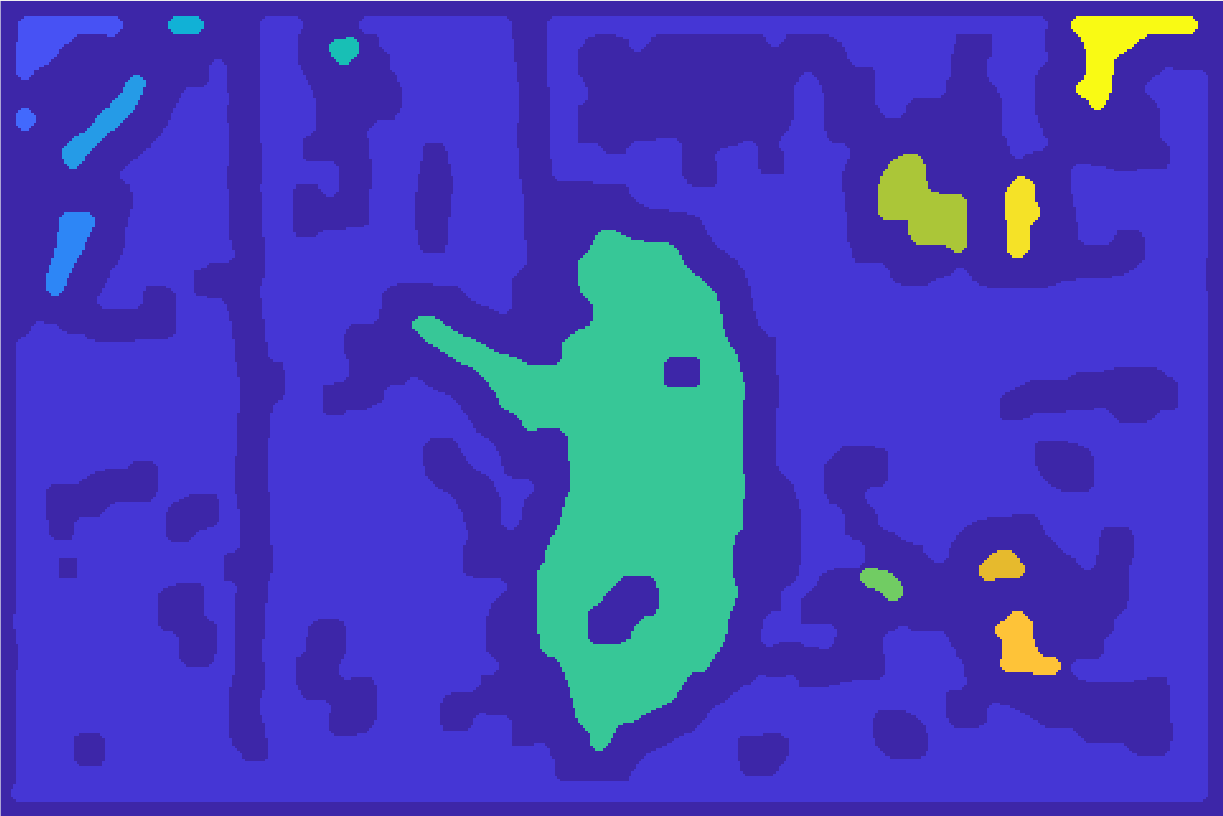}
  \caption{CCs, erosion at scale $R=7$}
\end{subfigure}
\begin{subfigure}[b]{0.14\textwidth}
  \includegraphics[width=\textwidth]{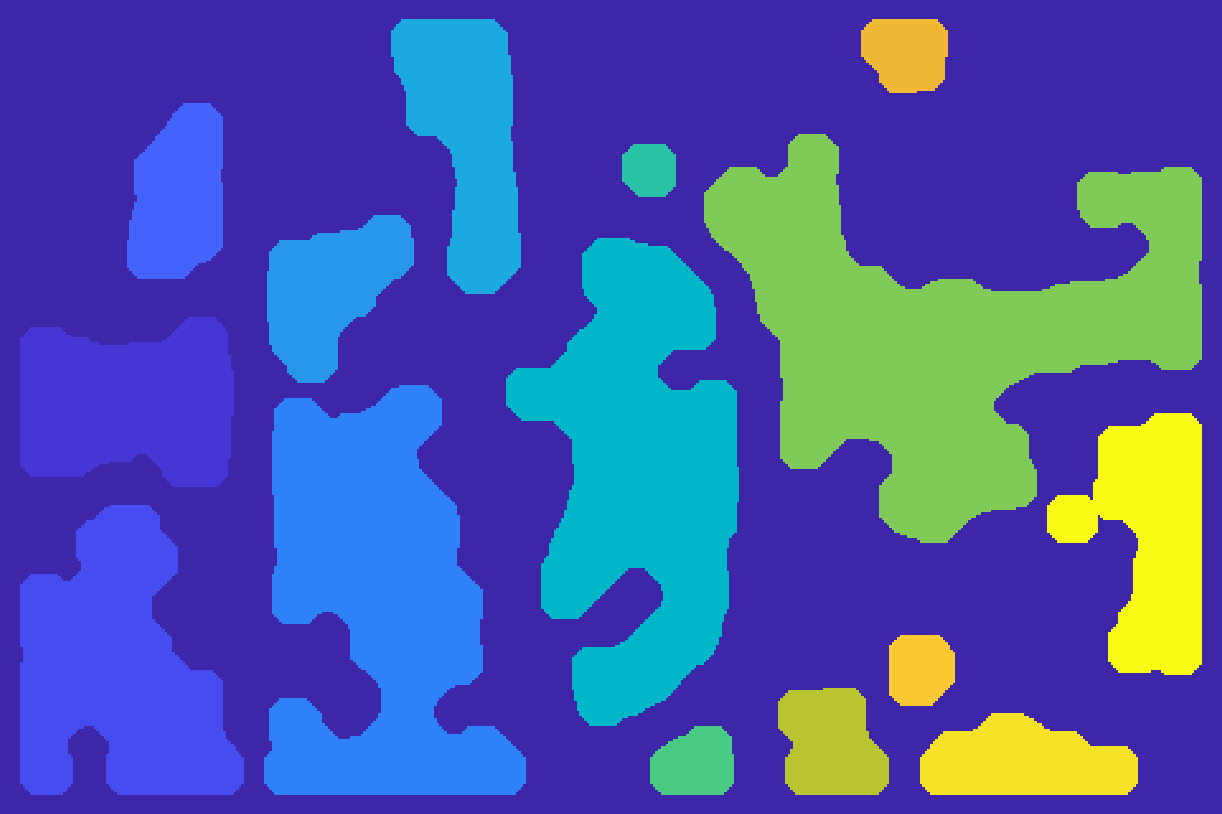}
  \caption{CCs, erosion at scale $R=15$}
\end{subfigure}
\begin{subfigure}[b]{0.14\textwidth}
  \includegraphics[width=\textwidth]{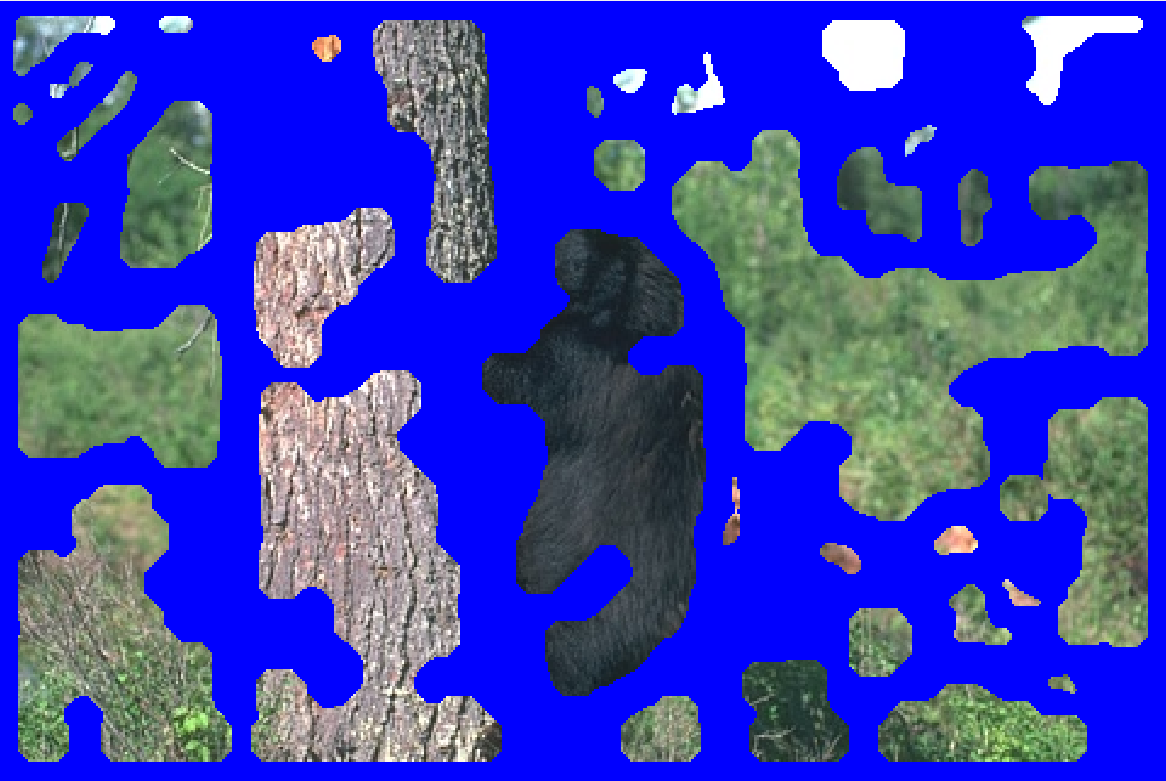}
  \caption{Initial seed regions}\label{subfig:initial-seeds}
\end{subfigure}

\caption{The estimated DT and the multi-scale erosions to obtain the initial seed regions. Notice how the multi-scale erosions make it possible to get both fine small regions as well as break up large regions containing false merges.}\label{fig:dwt}
\end{figure}

In practice, we found that the estimated seed regions sometimes contain false merges between segments where the detected contour was not completely closed. Thus, we use multi-scale erosions to remove some of these merges. We perform erosion with disks at multiple scales (different disk radius for each scale) and merge the non-overlapping CCs from different scales, starting from the largest erosion scale; see an example in Fig. \ref{fig:dwt}. We refer to this set of CCs as initial seed regions, denoted as $v_i$. 
Full details of the distance transform network are provided in the supplement.

\subsubsection{Seed merging}\label{merging-direct}

The initial seed regions are merged to larger seed regions as follows:

\begin{enumerate}
\item Specify a k-nearest-neighbors  graph $\mathcal{G}_s(V_s,E_s)$ where the vertices $V_s$ are the initial seed regions $v_i$.
\item Characterize each edge $e_{ij}$ with a feature vector $f_{ij}$ describing the seeds associated with the edge.
\item Use a learned classifier over $f_{ij}$ to assign a weight to each edge.
\item Threshold the weights on the edges to obtain a partition of $\mathcal{G}_s$ into connected components.
\item Specify the seed regions $s_i$ as the union of initial seed regions $v_i$ in each CC.
\end{enumerate}

%
%
The seed regions $s_i$ will serve as the seeds for the final segmentation stage (Sec. \ref{crf}). A threshold is thus associated with some segmentation and its corresponding precision and recall values. The different PR values for different thresholds compose the PR curve.

\textbf{The feature vector $f_{ij}$.} The constructed feature vector $f_{ij}$ should reflect whether the two seeds are likely to belong to the same segment if their average representations are similar, if they are spatially close, and if there are no significant edges between them.
This vector has two features. The first is the representation distance between the average representations of the connected nodes (seeds) $v_i, v_j$, denoted as $\mu_i, \mu_j$ and calculated by $f_{ij,1} = || \mu_i - \mu_j ||$.
The second, $f_{ij,2}$, is the geodesic distance between the two nodes $v_i$ and $v_j$ calculated, on an 8-connected graph where the nodes are the pixels and the weight on each edge is the edge strength. 
Several other features were explored, but did not improve the performance. 
%
%

\begin{figure}[t]
\centering
\includegraphics[width=0.45\textwidth]{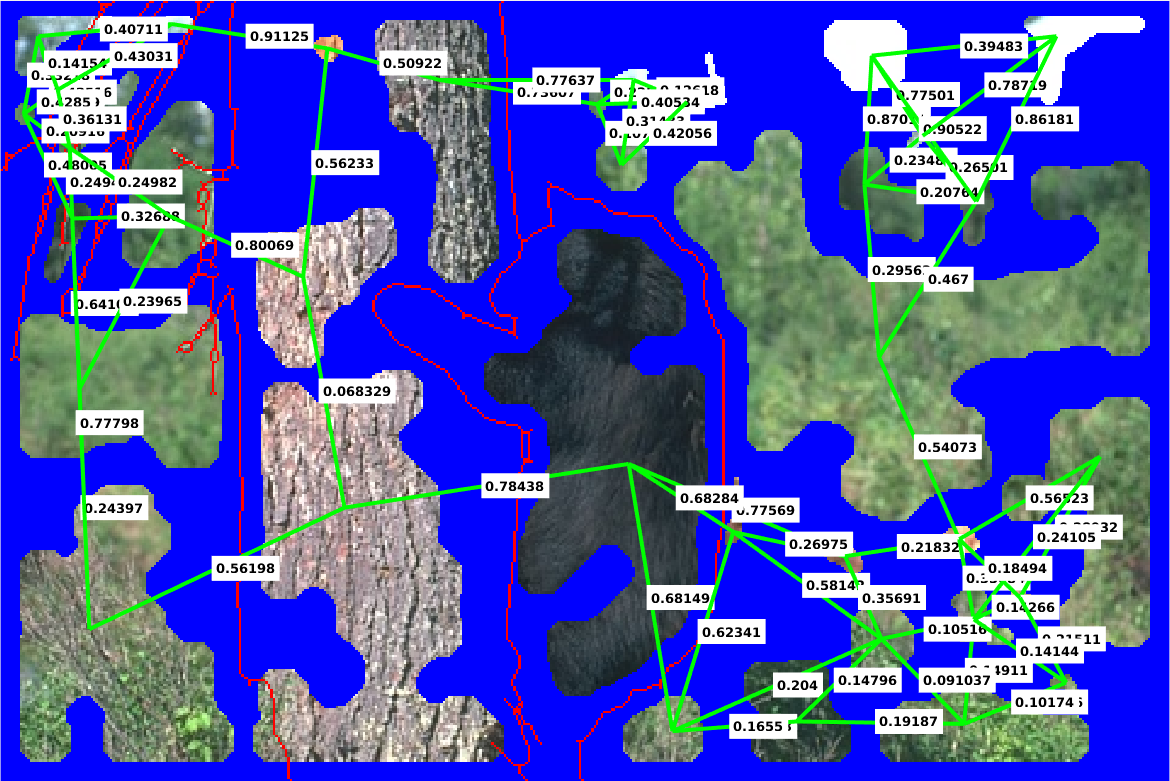}
\caption{The formed graph $\mathcal{G}_s(V_s,E_s)$ using the initial seed regions $v_i$ from Fig. \ref{subfig:initial-seeds}. The GT boundaries are marked in red. Each edge $e_{ij}$ is marked with a green line connecting two vertices $v_i, v_j$.}\label{fig:graph}
\end{figure}

\textbf{The learned classifier.} 
The weight on the edge $e_{ij}$ was set to be the classifier's soft output; see an example in Fig. \ref{fig:graph}. Classification results of the trained classifier on the test set are shown in Table \ref{table:seeds-alg-1}. We can see that the representation distance outperforms the geodesic distance when each is used alone. The combination of both achieves the best accuracy. 

\setlength{\tabcolsep}{4pt}
\begin{table}[!h]
\begin{center}
\begin{tabular}{lcc}
\hline\noalign{\smallskip}
Feature & Test accuracy \\
\noalign{\smallskip}
\hline
\noalign{\smallskip}
Geodesic distance & 78\% \\
Representation distance & 80.01\% \\
\textbf{Both} & \textbf{81.5\%} \\
\hline
\end{tabular}
\end{center}
\caption{Seed region classification results, alg. 1}\label{table:seeds-alg-1}
\end{table}
\setlength{\tabcolsep}{1.4pt}

For an alternative, indirect, seed generation process, see the supplementary material.

\subsection{Stage 2: Combining the representation and seed regions}\label{crf}

\subsubsection{Modeling pixel label probabilities}

The previous stage ends with a set of image regions $s_i$, each corresponding to a specific hypothesized segment and associated with a unique label. 
The next stage models a probability for each pixel to belong to each of the hypothesized segments, uses it  as the unary term in a conditional random field (CRF), and performs an estimated inference over the entire image.

To model the pixel-wise probability, we use a simple Gaussian modeling.
It is clear that a pixel is more likely to belong to an hypothesized segment if its representation is similar to the representation characterizing the segment, if it is spatially closer to the segment, and if there are no significant edges between the pixel and the corresponding seed region.
The likelihood of every pixel $p_i$, associated with a representation  $R_i$, relative to the $j$th hypothesized segment $s_j$, is modeled as: 
\begin{align}
  z_{ij} = C_{n} exp\left(-C_r D_r(p_i, s_j) - C_g D_g( p_i, s_j ) \right)
\end{align}

$D_r$ expresses the dissimilarity between the representations of the pixel and the hypothesized segment $s_j$ using the (squared) Mahalanobis distance:
\begin{align}
D_r(p_i, s_j) = (R_i - \mu_j)^T\Sigma_j^{-1} (R_i - \mu_j)
\end{align}

The representation characterizing the hypothesized segment $s_j$ is described by its mean, $\mu_j$, and by its covariance matrix $\Sigma_j$, modeled as a diagonal matrix. The distance $D_g(p_i, s_j)$ is the geodesic distance between the $i$th pixel and the $j$th segment. $C_{n}$ is a normalizing constant to make the expression a likelihood, which was not calculated in practice, and $C_r, C_g$ are learned parameters. These parameters were learned using grid search.

Then, to get a probability-like expression, we normalize each likelihood by the sum of all likelihood terms associated with all segments $j$. This normalization is done independently for each pixel and is equivalent to calculating the posterior probability, assuming equal priors.  
\begin{align}\label{Zij}
  Z_{ij} = \frac{z_{ij}}{\sum_{j} z_{ij}}.
\end{align}

\subsubsection{Segmentation using a conditional random field}

For the final stage, from which the final segmentation is obtained, we adopt the fully connected pairwise CRF (\cite{fc-crf}). In contrast to semantic segmentation, where the segments of interest are associated with specific categories taken from a finite set, neither the number of categories nor their identity is known. Thus, we cannot use a standard model-based NN to acquire the unary values (\cite{deeplab,crfrnn}), and this is where our representations come into play.

The unary values $\psi^u_{ij}$ are set as $\psi^u_{ij} = -log\left(Z_{ij}\right)$ according to our suggested probability-like expression from eq. \ref{Zij}
%
For the binary potentials we use the pairwise potentials as defined in \cite{fc-crf}. All CRF parameters were learned using grid search and inference was done as in \cite{fc-crf}.

To validate the necessity of a CRF later on, we also examine a segmentation obtained by setting the label of pixel $i$, $X_i$, as the most probable decision independently for every pixel, which can be considered a Bayesian-like decision $X_{i} = \argmax_j  \left\{ Z_{ij}\right\}$. We refer to this segmentation as \textit{DGS-unary} in our experiments.
%
%
%

\section{The Representation Learning Network}\label{repnet}
 
We learn the pixel-wise representation implicitly through our suggested classification task (sec. \ref{Learning-rep}).  We use a fully convolutional architecture based on a modified version of ResNet-50 \cite{resnet}, and make use of layers $conv1$ through $conv5\_3$. 

\begin{figure}[!h]
\centering
\includegraphics[width=0.45\textwidth]{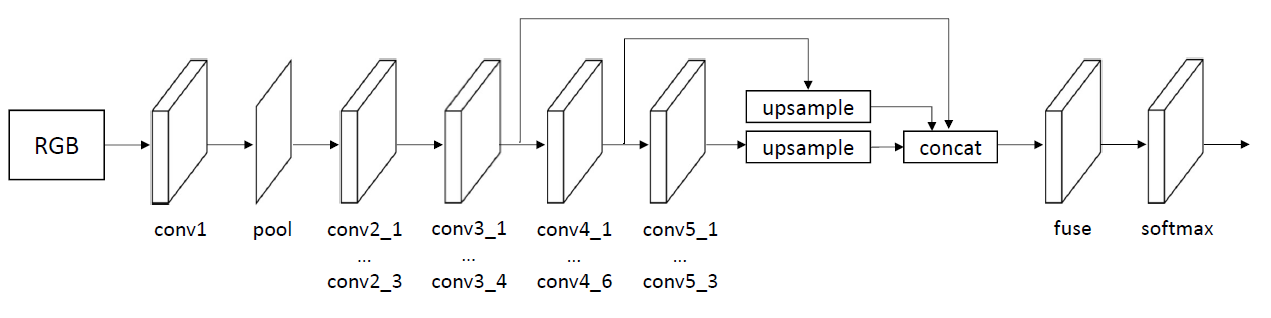}
\caption{Our suggested RepNet architecture.}
\label{repnet-arch}
\end{figure}

To increase the spatial output resolution, we adopt two common approaches: first, atrous (or dilated) convolutions \cite{deeplab} are used throughout layers $conv5\_x$. Second, we use skip connections of layers $conv3\_4, conv4\_6$ and concatenate them with layer $conv5\_3$, upsampling all to the resolution of $conv3\_4$, which is downsampled by $4$ compared to original input resolution. 
The concatenated layers pass through a final $fuse$ residual layer, to get a final feature depth of $512$ per pixel. An illustration of the network architecture (referred to as RepNet) is shown in Fig. \ref{repnet-arch}.

In general, the huge pixel-wise classification task (thousands of classes for each pixel) is a major hardware bottleneck that limits our resolution upsampling capabilities. While semantic segmentation task architectures are able to upsample to the original input resolution (the number of classes  are in the range of tens), we were limited to a resolution of $\frac{1}{4}\times$ of the original input image. The upsampling is done with bilinear interpolation. We found that here, deconvolution based upsampling did not improve performance.

The network was trained using a weighted cross-entropy loss. To improve segment separation, we increased the weight of the loss associated with pixels close to the boundary (closer than $d$) by a factor $w_b$.
The proposed pixel-wise representation is taken from the final layer before softmax, which we refer to as $fuse$.

\section{Experiments}
\label{results}

We first present the details of the representation learning procedure and the evaluation of these representations according to the experiment described in sec. \ref{validation-reps}. We then present quantitative and qualitative segmentation results.

\begin{figure*}[h]
\centering

\begin{subfigure}[b]{0.206\textwidth}
  \includegraphics[width=\textwidth]{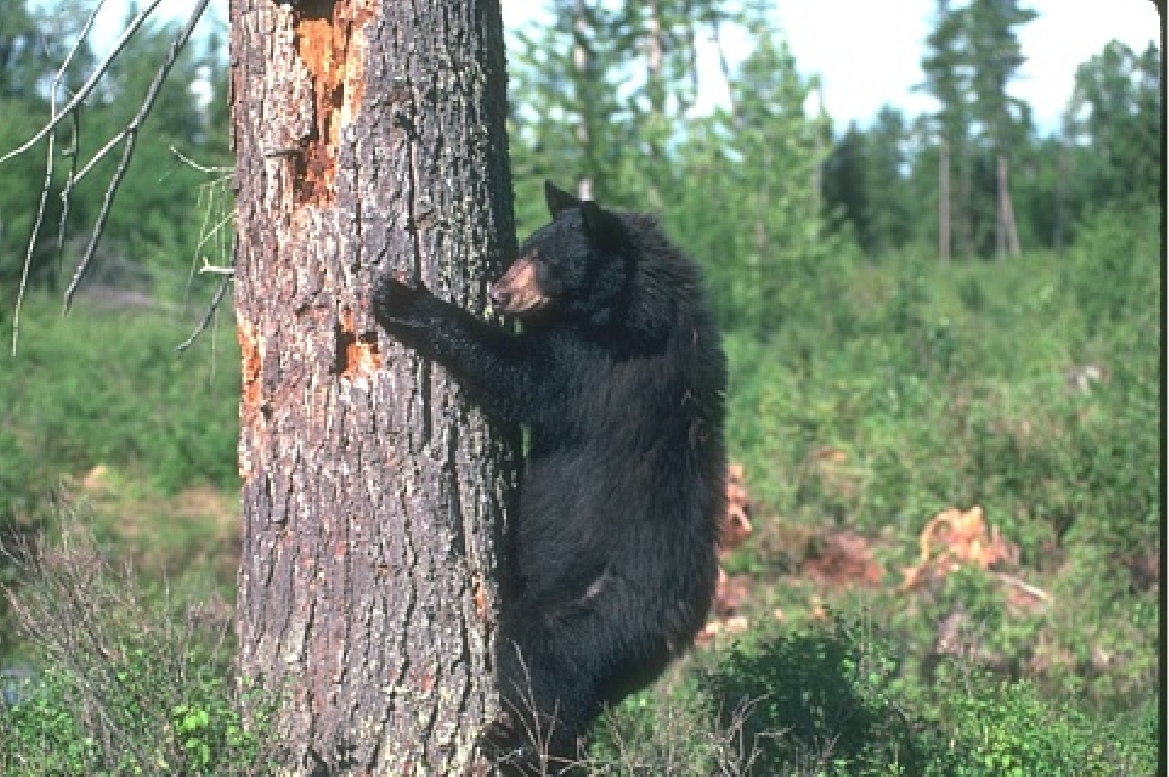}
\end{subfigure}
\begin{subfigure}[b]{0.206\textwidth}
  \includegraphics[width=\textwidth]{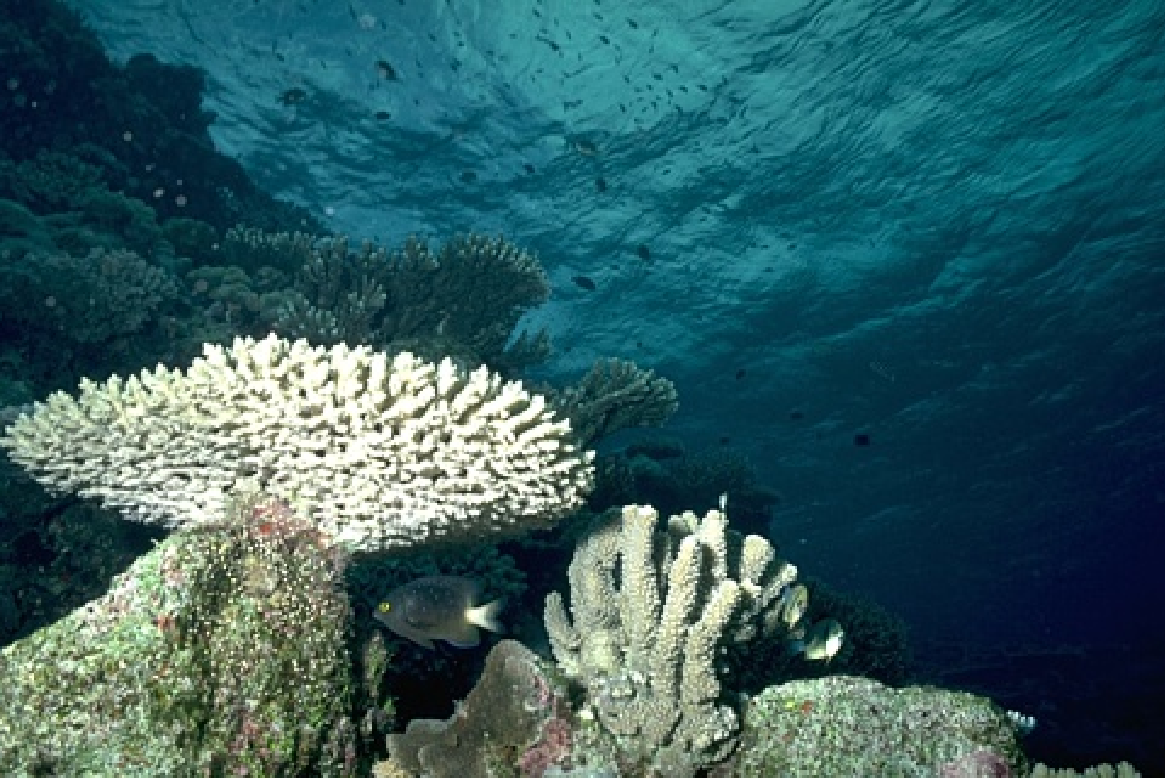}
\end{subfigure}
\begin{subfigure}[b]{0.206\textwidth}
  \includegraphics[width=\textwidth]{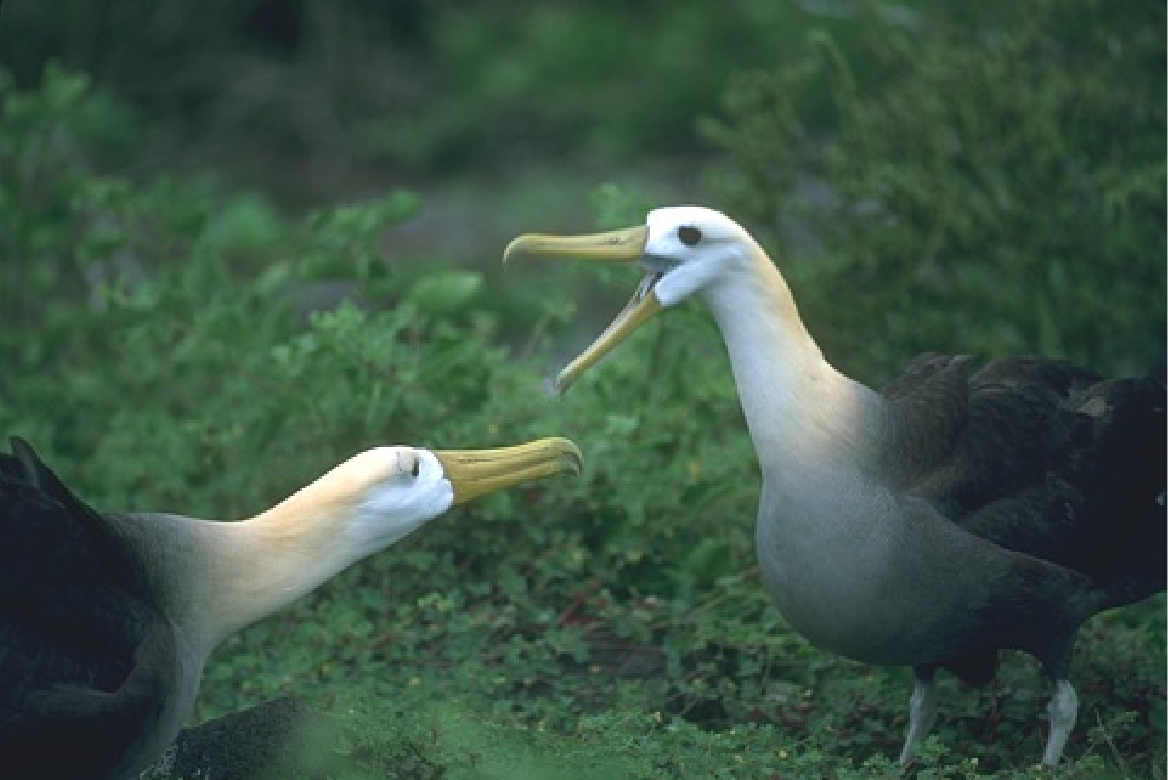}
\end{subfigure}
\begin{subfigure}[b]{0.206\textwidth}
  \includegraphics[width=\textwidth]{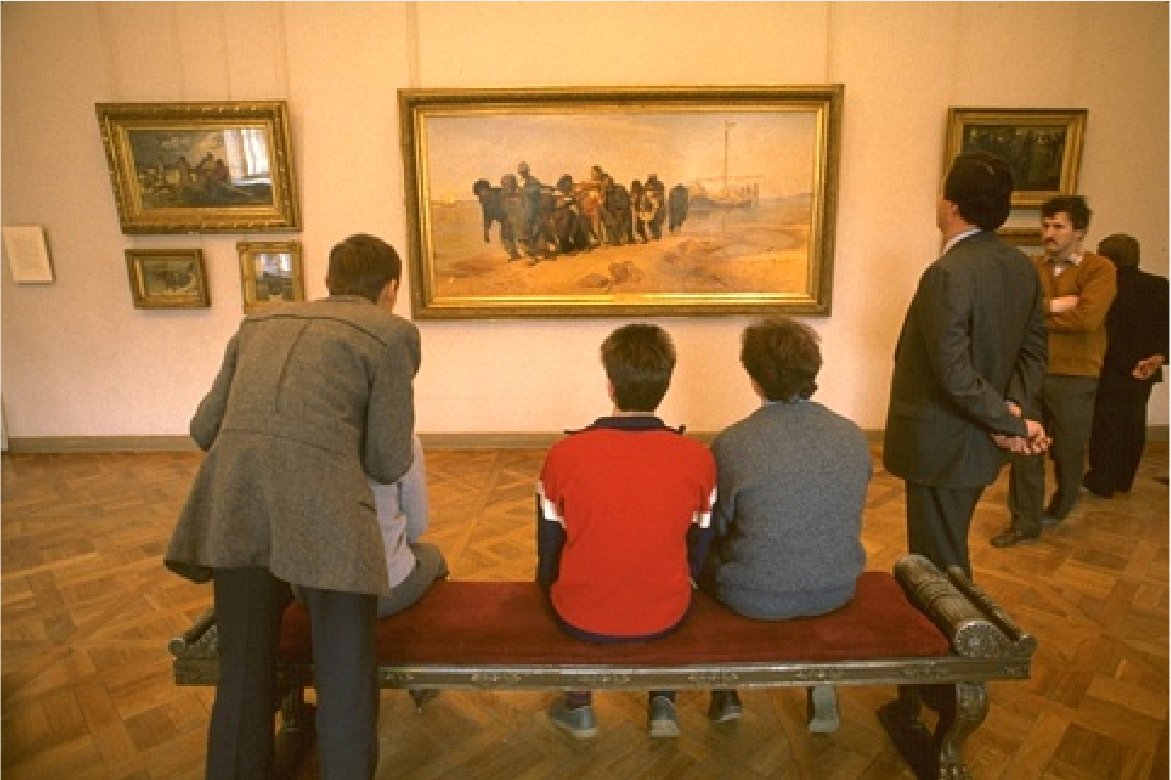}
\end{subfigure}
\vspace{2 mm}

\begin{subfigure}[b]{0.2055\textwidth}
  \includegraphics[width=\textwidth]{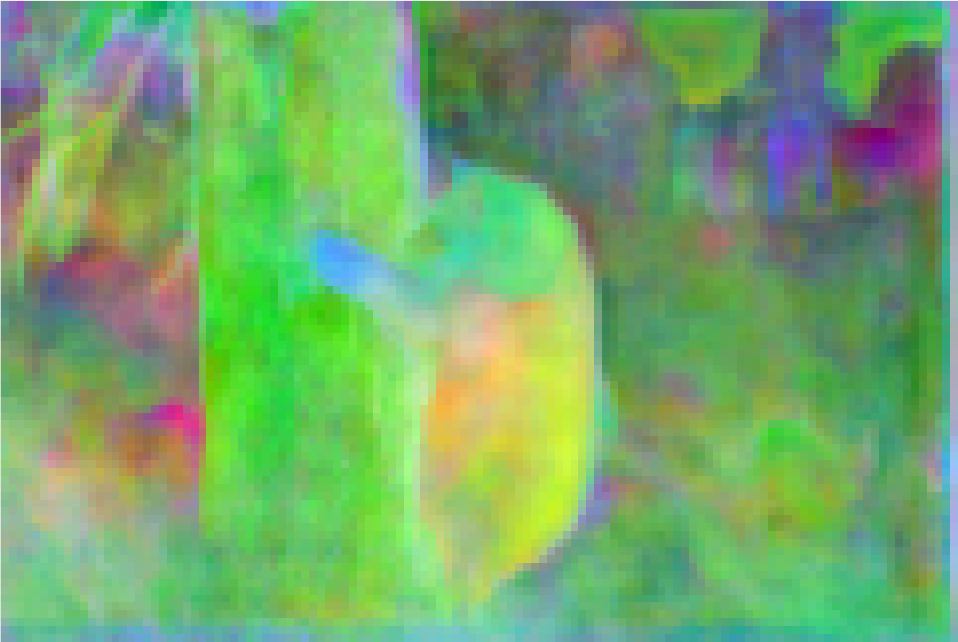}
\end{subfigure}
\begin{subfigure}[b]{0.2055\textwidth}
  \includegraphics[width=\textwidth]{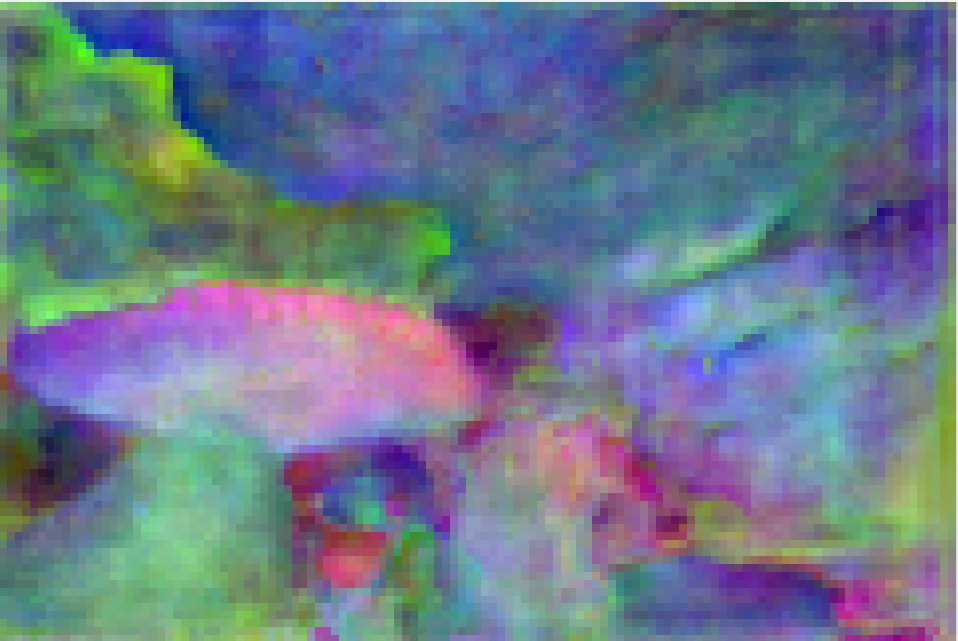}
\end{subfigure}
\begin{subfigure}[b]{0.2055\textwidth}
  \includegraphics[width=\textwidth]{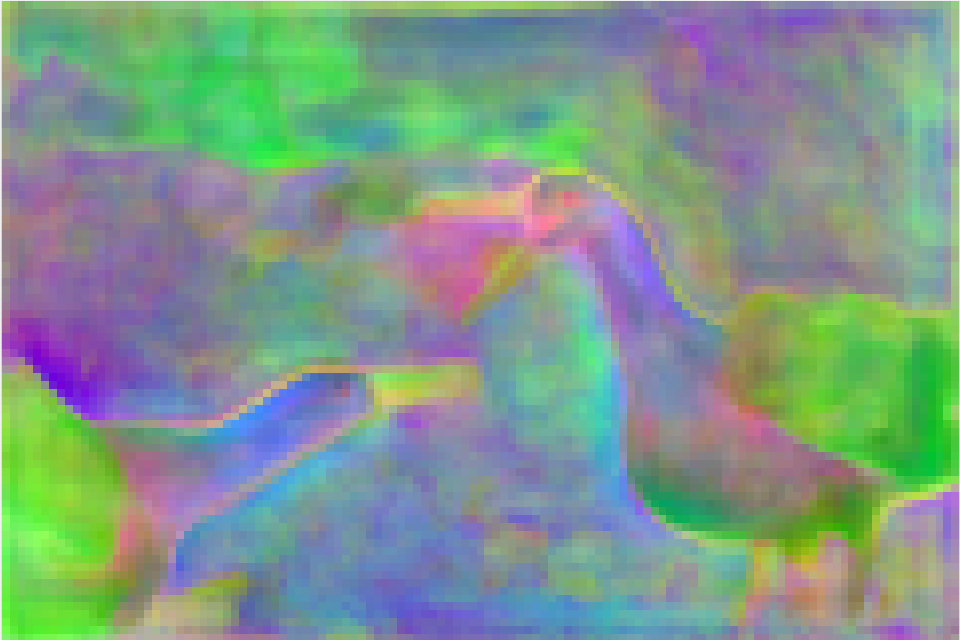}
\end{subfigure}
\begin{subfigure}[b]{0.2055\textwidth}
  \includegraphics[width=\textwidth]{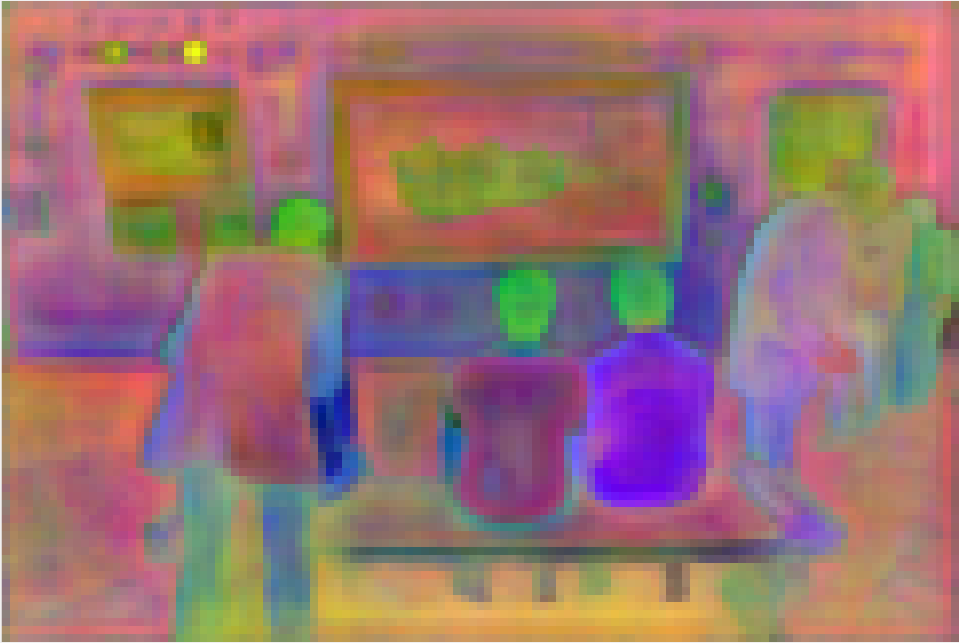}
\end{subfigure}
\vspace{2 mm}

\begin{subfigure}[b]{0.2055\textwidth}
  \includegraphics[width=\textwidth]{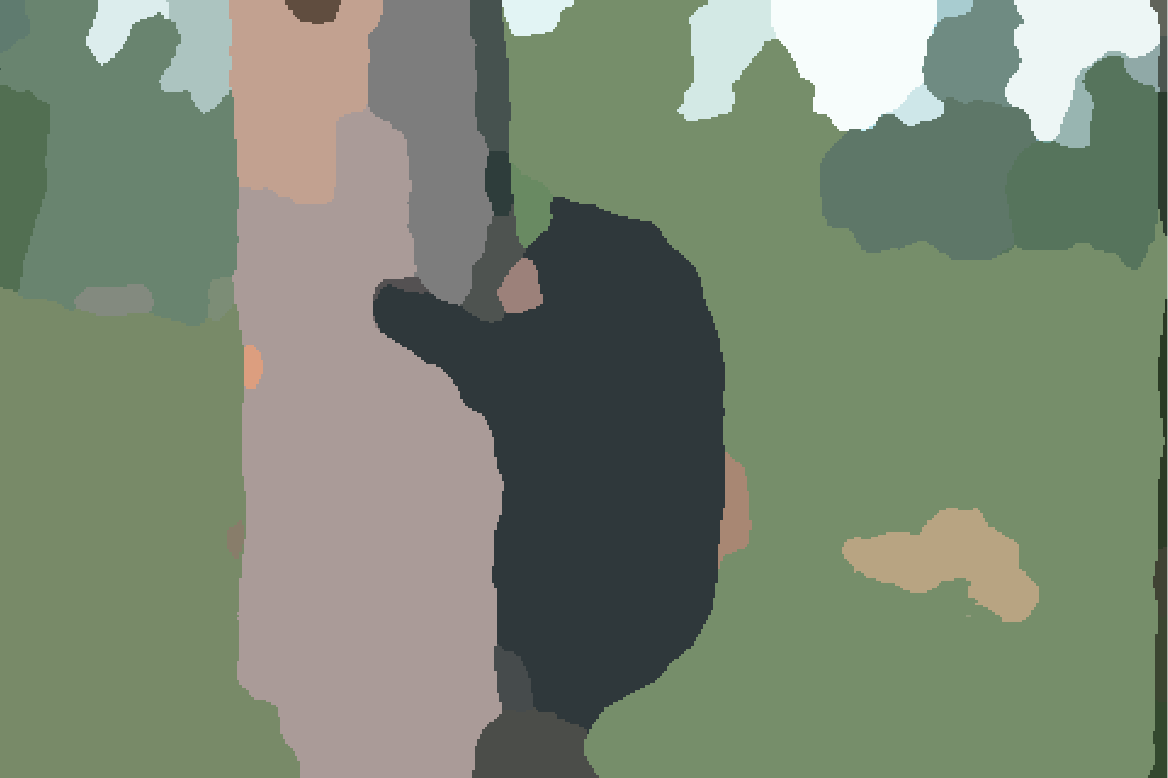}
\end{subfigure}
\begin{subfigure}[b]{0.2055\textwidth}
  \includegraphics[width=\textwidth]{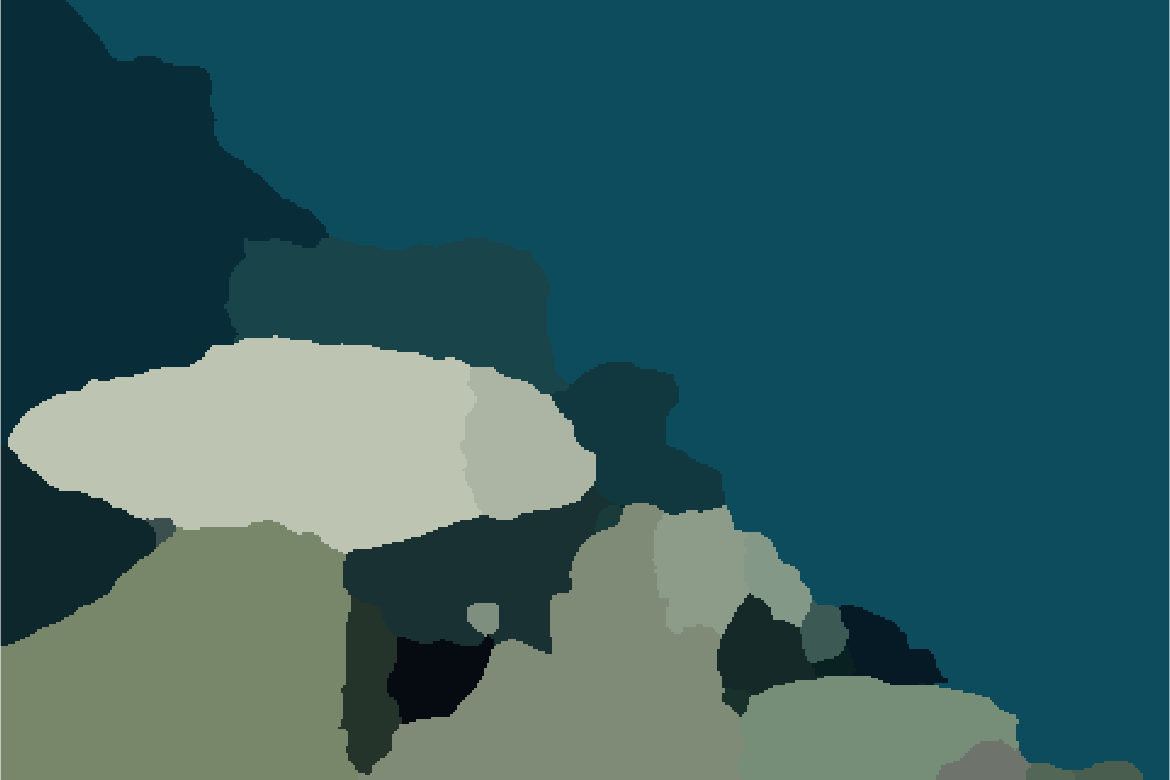}
\end{subfigure}
\begin{subfigure}[b]{0.2055\textwidth}
  \includegraphics[width=\textwidth]{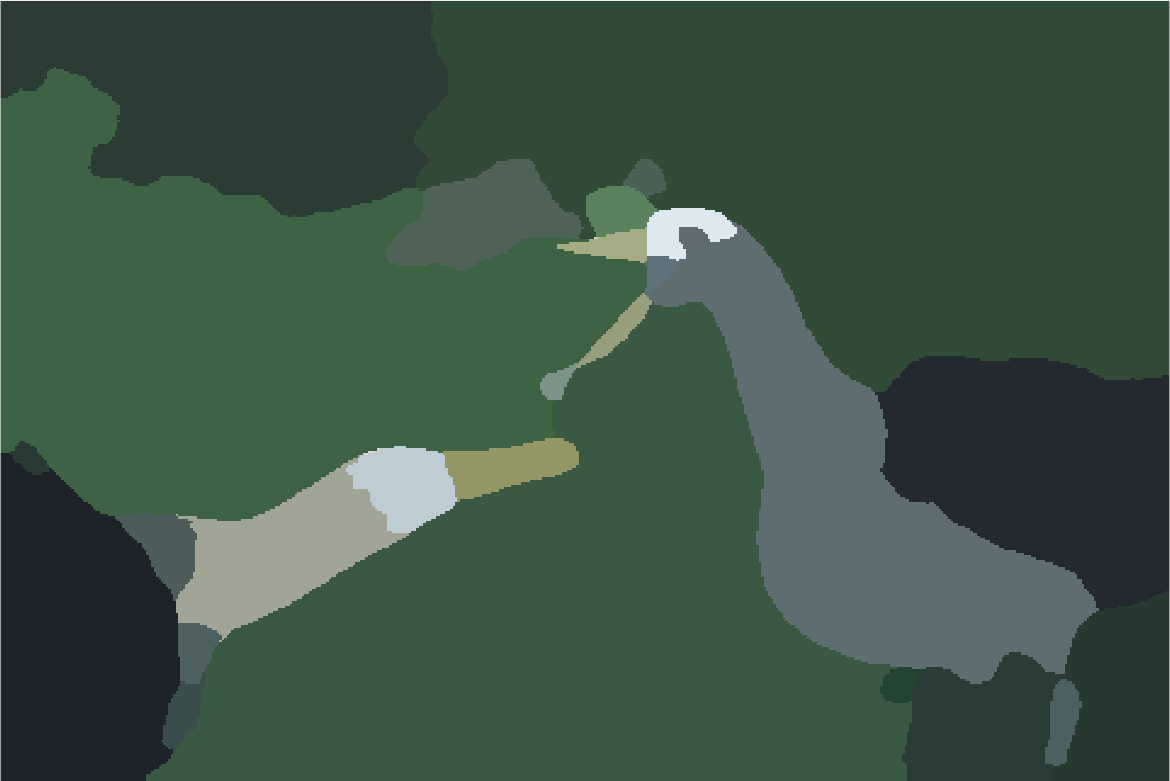}
\end{subfigure}
\begin{subfigure}[b]{0.2055\textwidth}
  \includegraphics[width=\textwidth]{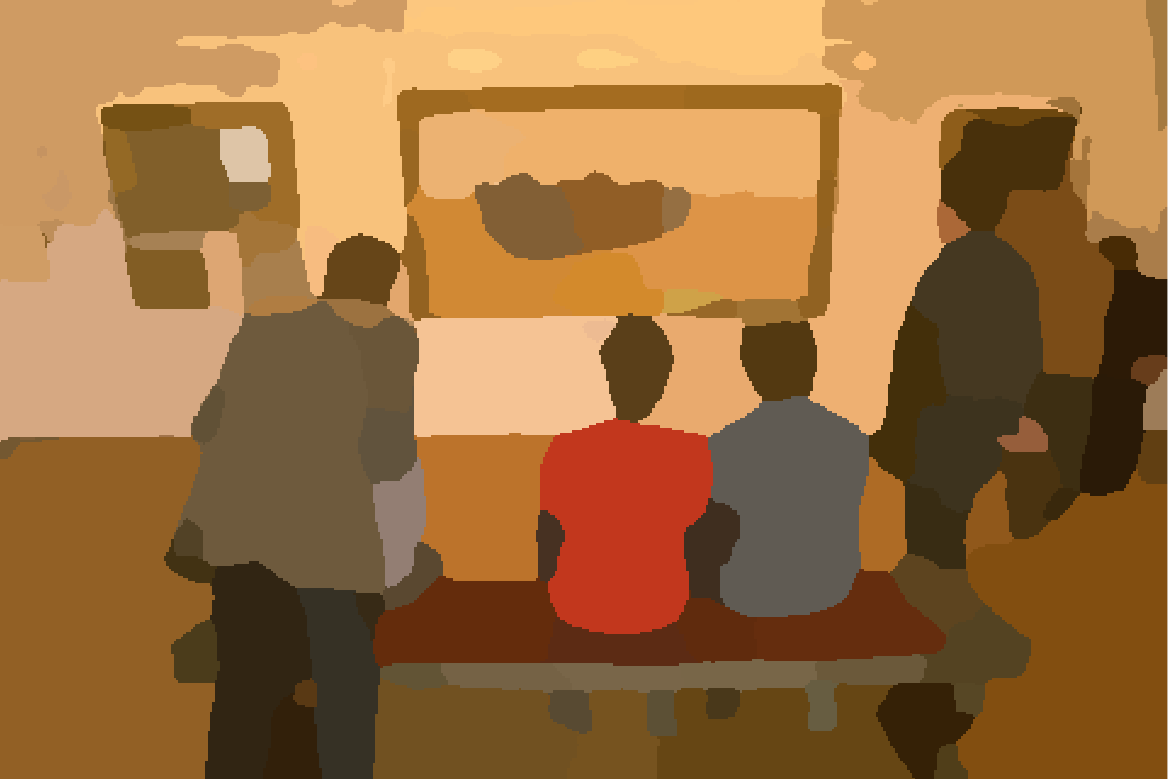}
\end{subfigure}
\vspace{2 mm}

\begin{subfigure}[b]{0.2055\textwidth}
  \includegraphics[width=\textwidth]{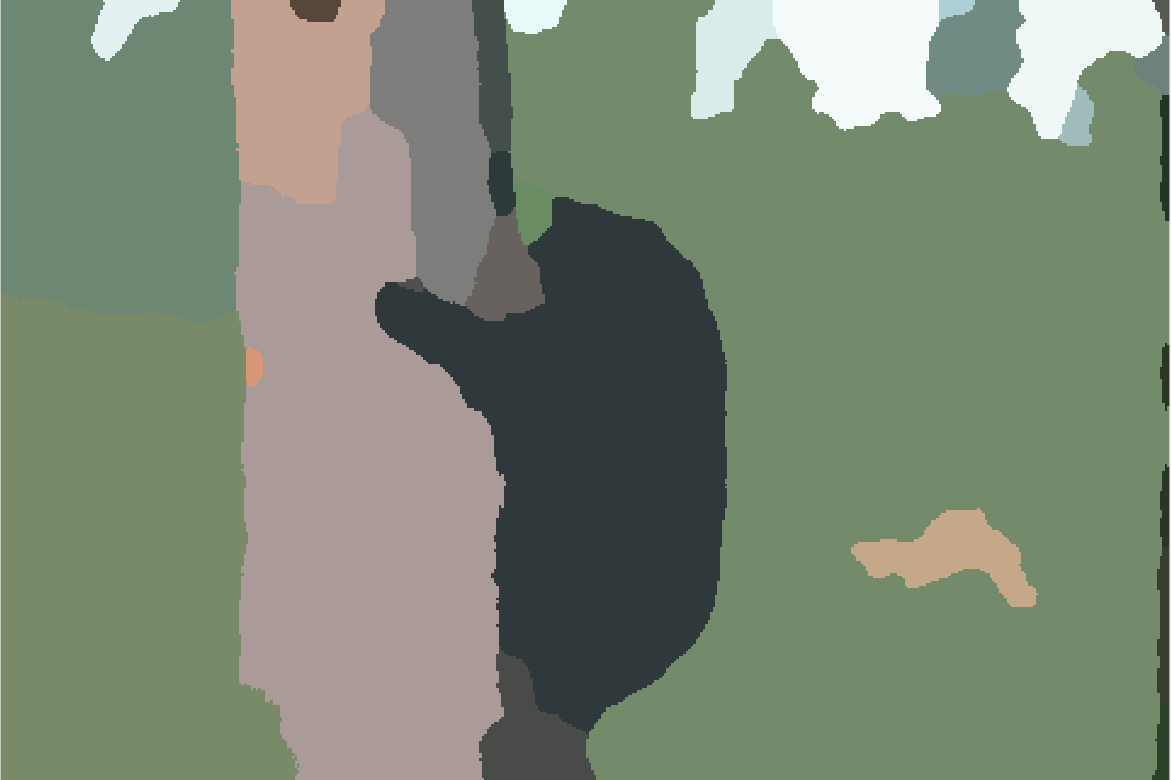}
\end{subfigure}
\begin{subfigure}[b]{0.2055\textwidth}
  \includegraphics[width=\textwidth]{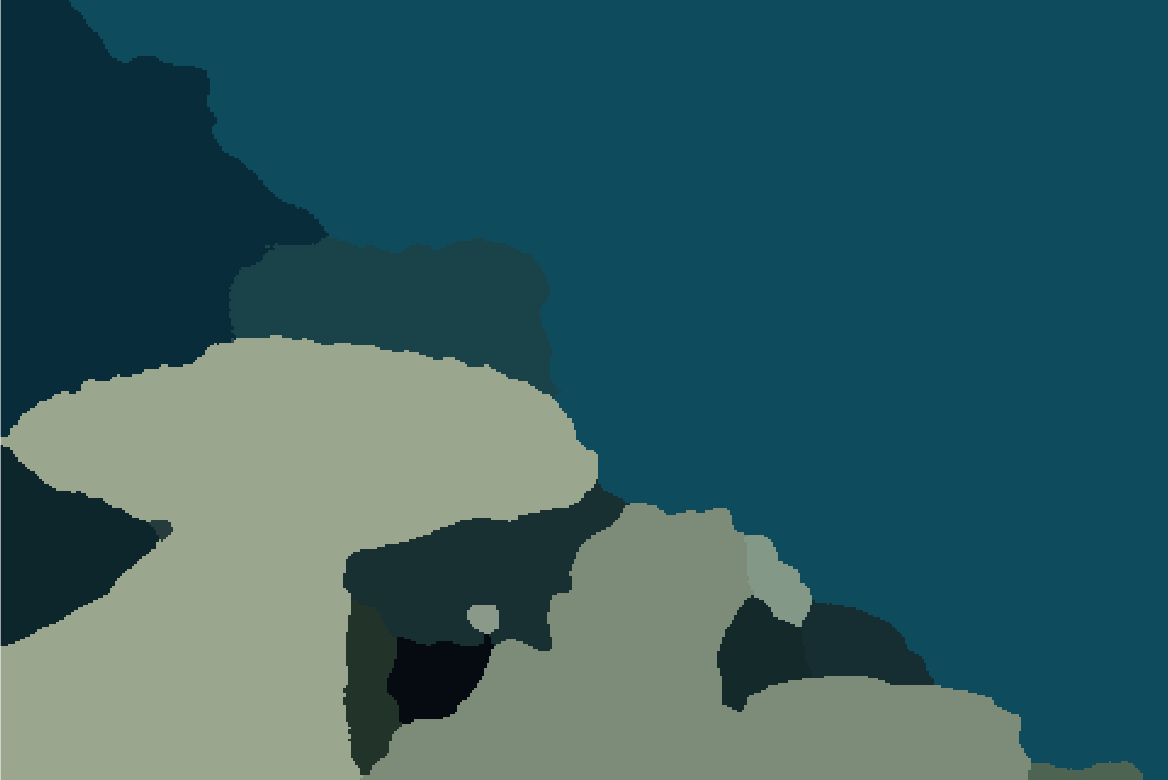}
\end{subfigure}
\begin{subfigure}[b]{0.2055\textwidth}
  \includegraphics[width=\textwidth]{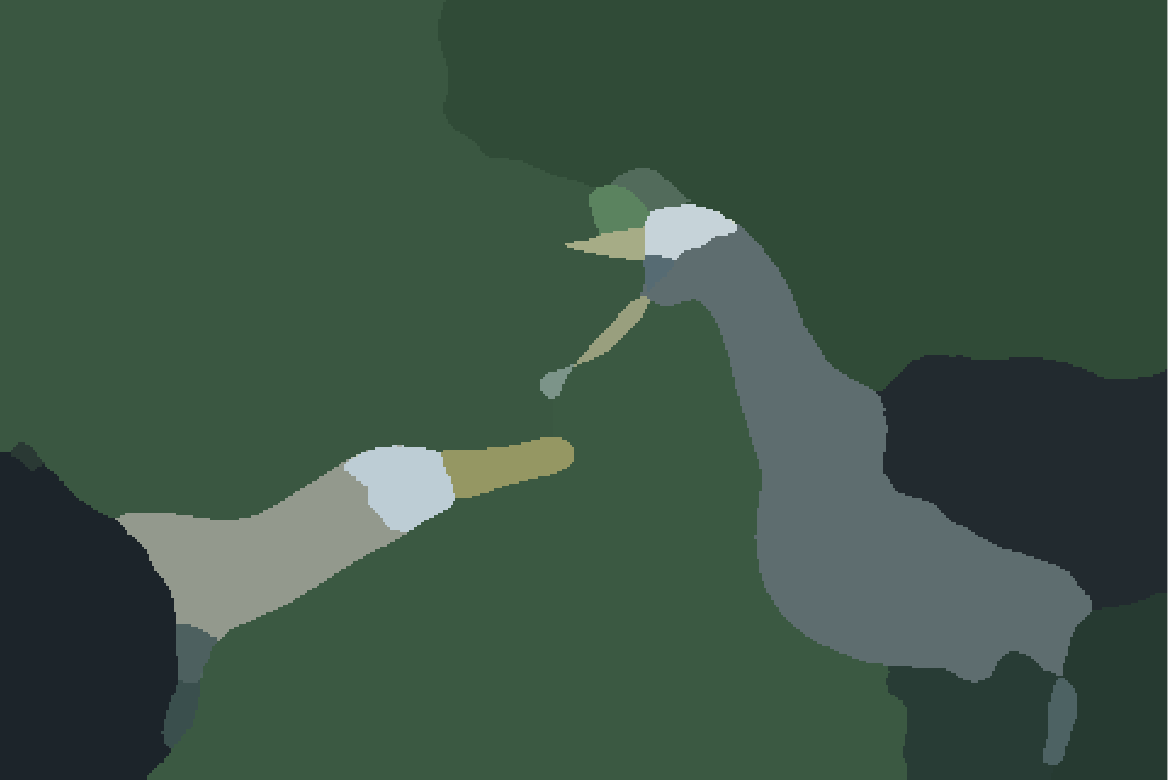}
\end{subfigure}
\begin{subfigure}[b]{0.2055\textwidth}
  \includegraphics[width=\textwidth]{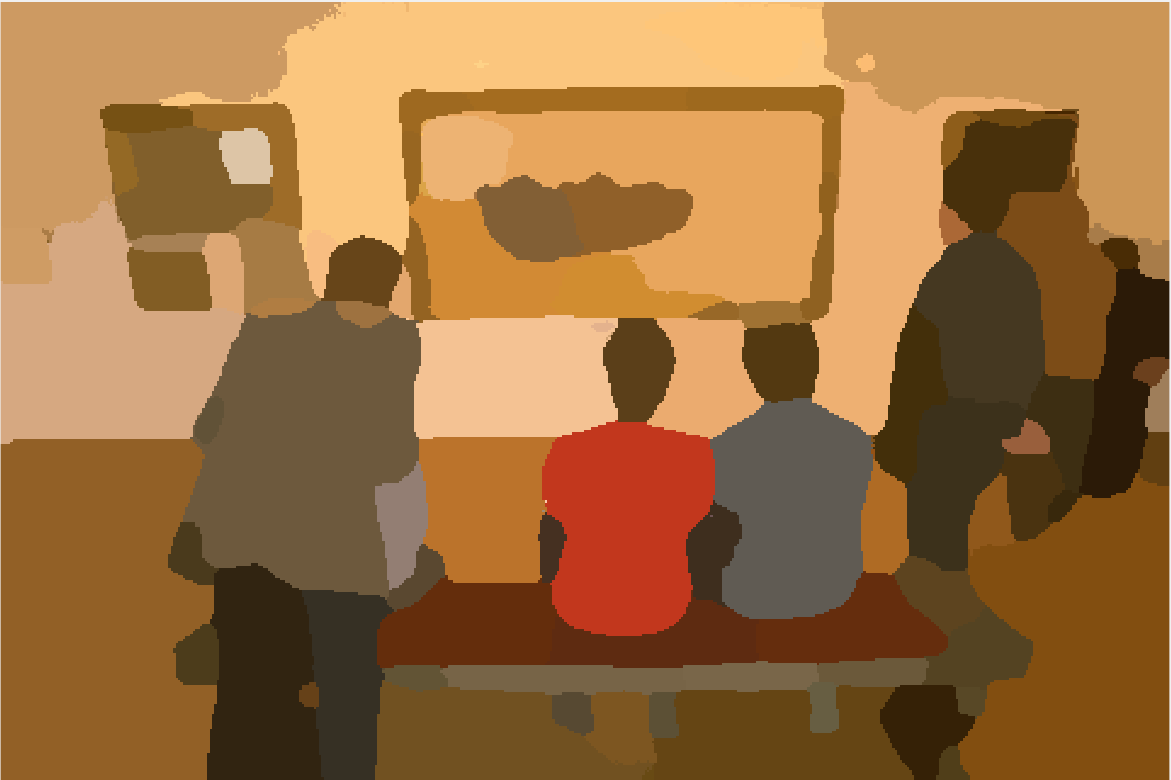}
\end{subfigure}
\vspace{2 mm}

\begin{subfigure}[b]{0.2055\textwidth}
  \includegraphics[width=\textwidth]{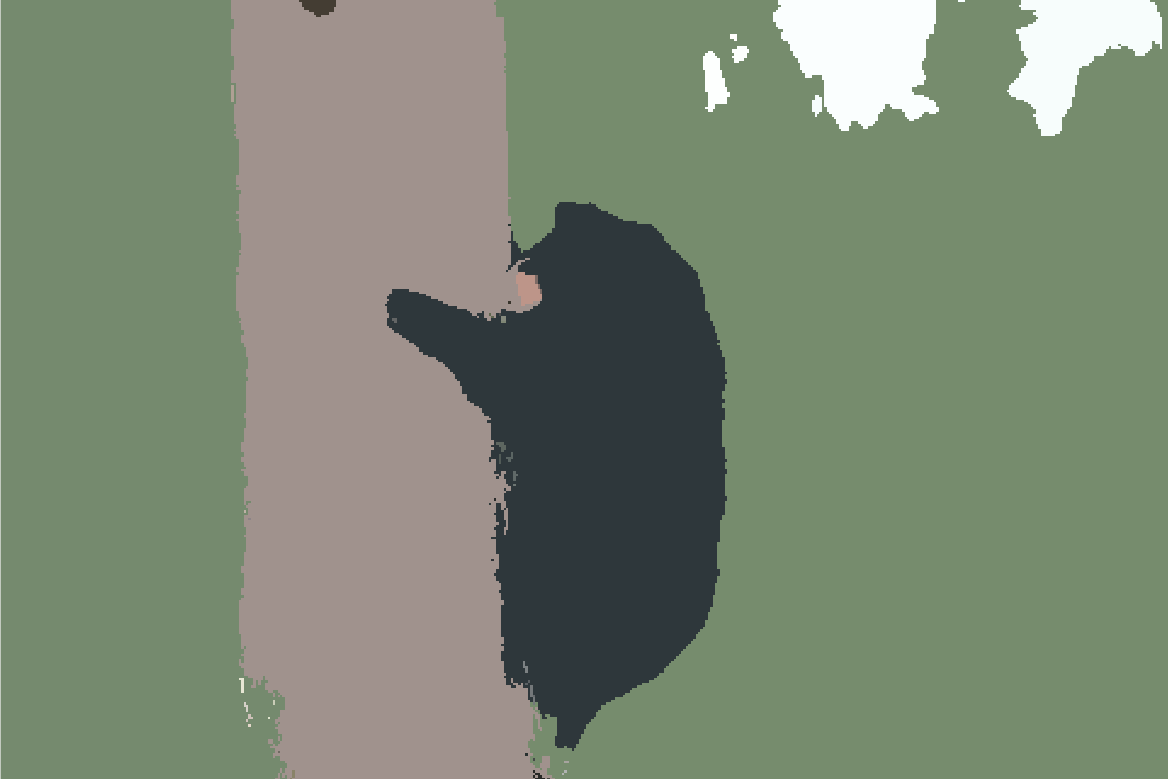}
\end{subfigure}
\begin{subfigure}[b]{0.2055\textwidth}
  \includegraphics[width=\textwidth]{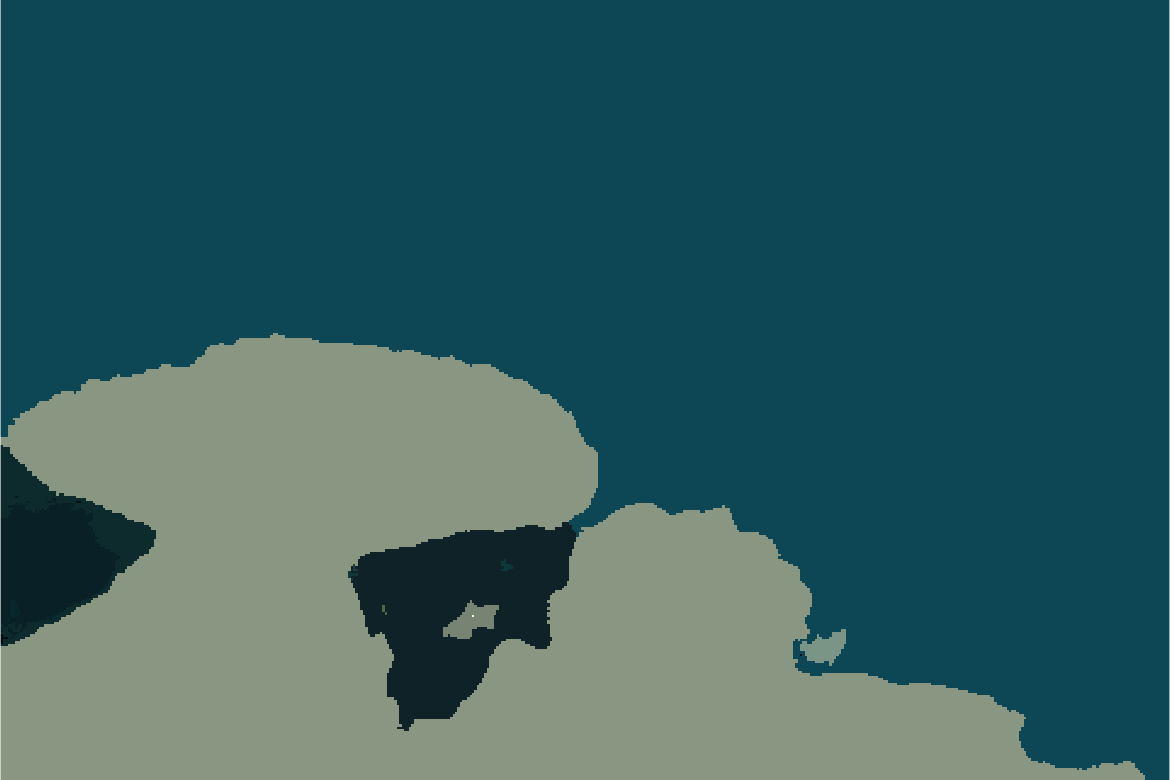}
\end{subfigure}
\begin{subfigure}[b]{0.2055\textwidth}
  \includegraphics[width=\textwidth]{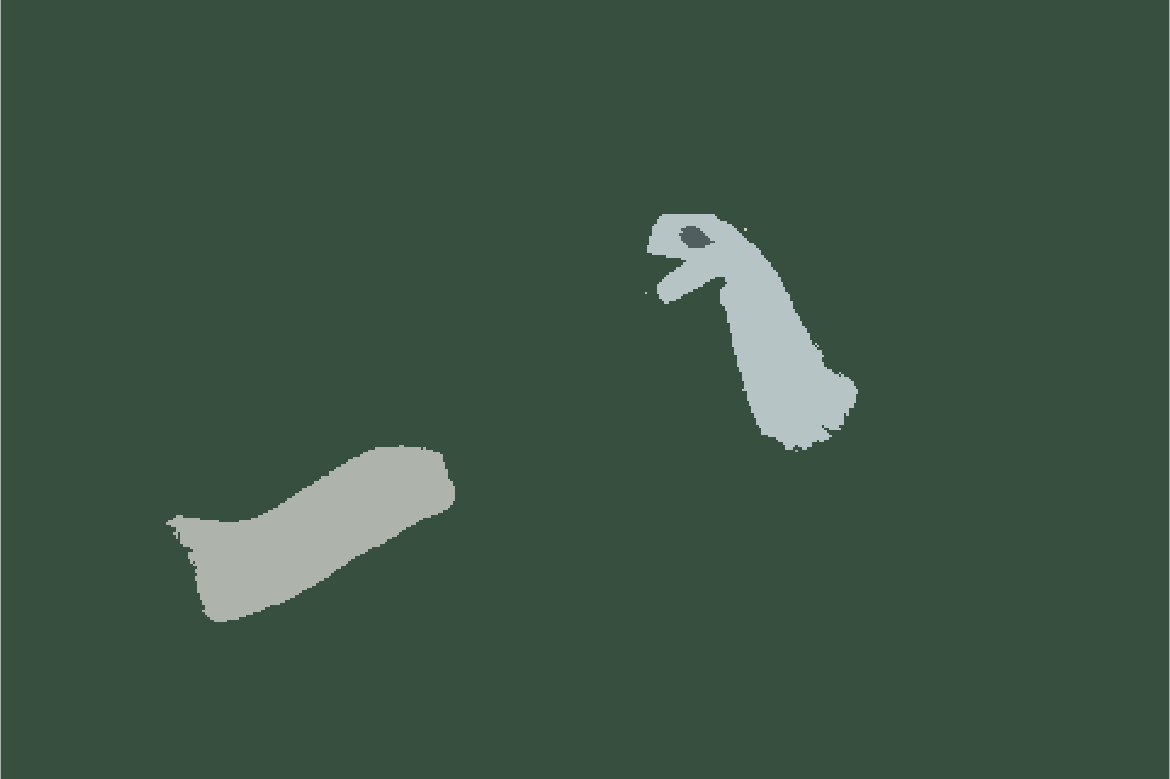}
\end{subfigure}
\begin{subfigure}[b]{0.2055\textwidth}
  \includegraphics[width=\textwidth]{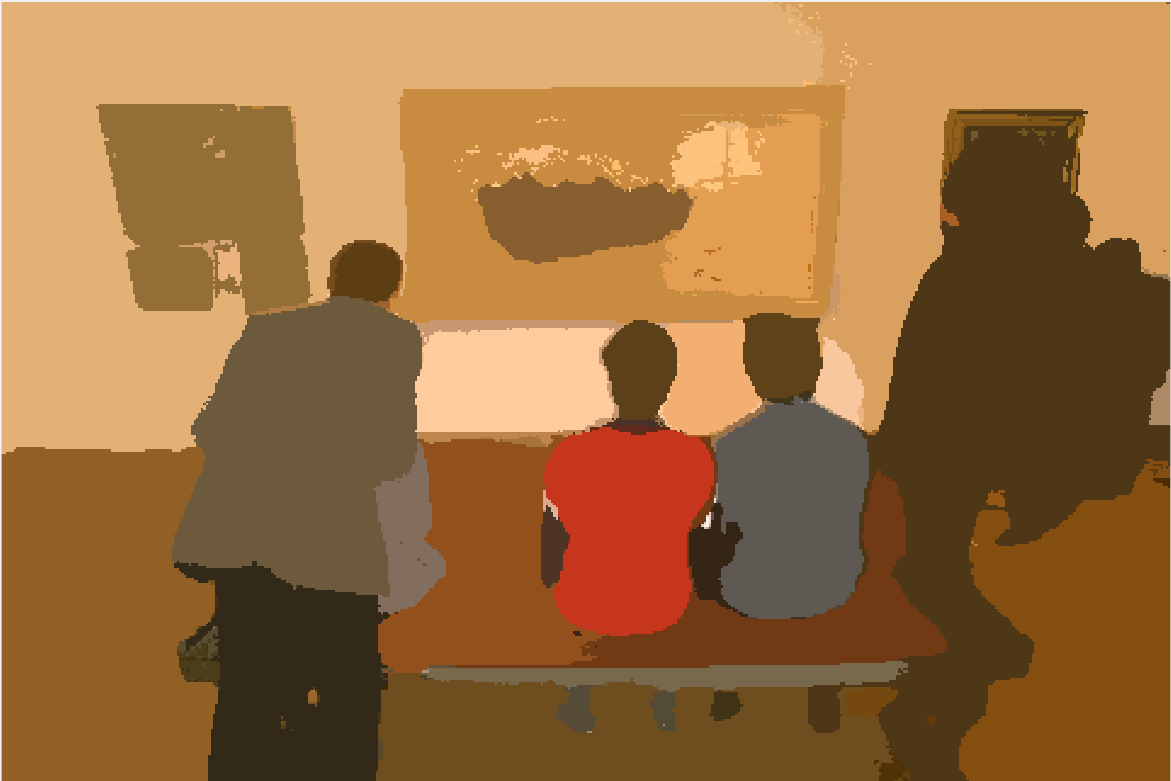}
\end{subfigure}

\caption{Segmentation results on BSDS500. In each column, from top to bottom: the image; representation space virtual colors; segmentation with DGS-unary; segmentation with DGS-SPTs; segmentation with DGS-DT.}\label{fig:bsds_segs}
\end{figure*}

\subsection{Representation learning}\label{rep-results}

We trained the representation network over the classification task described in sec. \ref{implicit}. We started with ImageNet pre-trained weights, and trained it first with 1000 images containing $5540$ segments from the Pascal Context dataset (\cite{pascal_context}), and then with $300$ \textit{trainval} images ($2060$ segments) from the BSDS dataset. We began with a learning rate of $0.001$ and achieved $95.2\%$ training accuracy.

Table \ref{table:pixels} shows the results of the pixel classification task from sec. \ref{validation-reps}. We compared the representations obtained from our implicit learning method (sec. \ref{implicit}) with representations learned as follows: triplet loss, representation from a network trained for semantic segmentation (\cite{deeplab}), and representations from a network trained for material classification 
(\cite{minc}). We also compare with the following pixel representations: RGB, L*a*b and Gabor filters.
Clearly, our proposed representation achieves the best result. 

\setlength{\tabcolsep}{4pt}
\begin{table}[!h]
\begin{center}
\begin{tabular}{lcc}
\hline\noalign{\smallskip}
Representation & Test accuracy \\
\noalign{\smallskip}
\hline
\noalign{\smallskip}
Gabor filters & 56.09\% \\
RGB & 57.67\% \\
L*a*b & 58.25\% \\
Material classification net \cite{minc} & 70.14\% \\
DeepLab \cite{deeplab} & 71.94\% \\
Triplet loss & 76.34\% \\
Ours (implicit learning) & \textbf{81.04\%} \\
\hline
\end{tabular}
\end{center}
\caption{Pixel pair classification results}\label{table:pixels}
\end{table}
\setlength{\tabcolsep}{1.4pt}

\subsection{Generic segmentation}

We ran DGS and optimized its parameters using grid search. The optimal parameters were: $w^{(1)} = 6, w^{(2)} = 1, C_r = 1.25, C_g = 0.5$. For the CRF we used  $\theta_a = 60, \theta_b = 10, \theta_{\gamma} = 3$. The results are compared with other algorithms on the $F_b$ and $F_{op}$ measures; see numerical results in Table \ref{table:bsds}, qualitative results in Fig. \ref{fig:bsds_segs} and PR curves in the supplement. DGS-DT denotes the algorithm in sec. \ref{alg1} which relies on the estimated DT. DGS-SPTs refers to the alternative algorithm from the supplement. DGS-unary refers to a segmentation without CRF processing. Note that both parameters $C_r, C_g$ are non-zero, implying that the best  results for our algorithm are obtained when the information from both components is combined. 
Our algorithm did not achieve the state-of-the-art results. 

\section{Conclusion}
\label{chap:conclusion}

We proposed a new approach to generic image segmentation that leverages the strengths of DNNs through pixel-wise representations. The representations are learned through a formulation of a supervised learning task that better suits our goal. These representations obtained state-of-the-art pixel similarity scores, serving as evidence that they capture characteristics that distinguish between different segments and suggesting that our approach generalizes well for segments not seen in the training set. The use of these representations through several stages of our proposed algorithm are promising evidence of their advantages for generic segmentation. Further work is required to achieve optimal results.

\setlength{\tabcolsep}{3pt}
\begin{table}[!h]
\small
\begin{center}
\begin{tabular}{lccc|ccc}
\hline\noalign{\smallskip}
&& $F_b$ & & & $F_{op}$ & \\
Method & ODS & OIS & AP & ODS & OIS & AP\\
\noalign{\smallskip}
\hline
\noalign{\smallskip}
DGS-unary (Ours) & 0.715 & 0.739 & 0.703 & 0.308 & 0.341 & 0.229\\
DGS-SPTs (Ours) & 0.727 & 0.749 & 0.732 & 0.313 & 0.350 & 0.237\\
DGS-DT (Ours) & 0.666 & 0.699 & 0.559) & 0.347 & 0.371 & 0.232\\
COB \cite{cob} & 0.793 & 0.820 & 0.859 & - & - & -\\
HED \cite{hed} & 0.780 & 0.796 & 0.834 & 0.415 & 0.466 & 0.333\\
LEP \cite{lep} & 0.757 & 0.793 & 0.828 & 0.417 & 0.468 & 0.334\\
MCG \cite{mcg} & 0.747 & 0.779 & 0.759 & 0.380 & 0.433 & 0.271\\
gPb-UCM \cite{ucm} & 0.726 & 0.760 & 0.727 & 0.348 & 0.385 & 0.235\\
Mshift \cite{meanshift} & 0.601 & 0.644 & 0.493 & 0.229 & 0.292 & 0.122\\
Ncut \cite{ncuts} & 0.641 & 0.674 & 0.447 & 0.213 & 0.270 & 0.096\\
\noalign{\smallskip}
\hline
\end{tabular}
\end{center}
\caption{BSDS500 evaluation results summary}\label{table:bsds}
\end{table}

\vspace{1cm}
{\small
\bibliographystyle{ieee}
\bibliography{egbib}
}

\end{document}